\documentclass{article}

\usepackage[colorinlistoftodos]{todonotes}

\usepackage{microtype}
\usepackage{graphicx}
\usepackage{subfigure}
\usepackage{booktabs} % for professional tables
\usepackage{wrapfig}

\usepackage{hyperref}
\hypersetup{hidelinks}
\usepackage{float}

\usepackage[preprint]{neurips_2026}

\usepackage{amsmath}
\usepackage{amssymb}
\usepackage{mathtools}
\usepackage{amsthm}

\usepackage[capitalize,noabbrev]{cleveref}

\theoremstyle{plain}

\theoremstyle{definition}

\theoremstyle{remark}

\usepackage[most]{tcolorbox}
\usepackage{listings}
\usepackage[table]{xcolor}
\usepackage{makecell}
\usepackage{enumitem}

\newtcblisting{PromptBox}[1]{
enhanced,
listing only,
colback=gray!3,
colframe=black!75,
boxrule=0.5pt,
arc=2pt,
left=8pt,
right=8pt,
top=10pt,
bottom=10pt,
fonttitle=\bfseries\sffamily\small,
colbacktitle=white,
coltitle=black,
title={#1},
attach boxed title to top left={yshift=-2mm, xshift=3mm},
boxed title style={sharp corners, boxrule=0.5pt},
breakable,
listing options={
basicstyle=\small\ttfamily,
breaklines=true,
columns=fullflexible,
keepspaces=true,
escapeinside={(*}{*)}
}
}

\newcommand{\hilite}[1]{\textcolor{red}{#1}}
\newcommand{\note}[1]{\textcolor{blue}{\textbf{\texttt{[#1]}}}}

\renewcommand{\hilite}[1]{#1}
\renewcommand{\note}[1]{}

\title{
Is This Your Final Answer?\\
Cross-Contextual Consistency as a Measure of LLM Credibility
}

\author{%
Siyang Wu \\
Data Science Institute \\
University of Chicago \\
Chicago, USA \\
\texttt{siyangwu@uchicago.edu}
\And
Yibo Jiang \\
Department of Computer Science \\
University of Chicago \\
Chicago, USA \\
\texttt{yiboj@uchicago.edu}
\And
Bryon Aragam \\
Booth School of Business \\
University of Chicago \\
Chicago, USA \\
\texttt{bryon@chicagobooth.edu}
}

\begin{document}

\maketitle

\begin{abstract}
Large language models (LLMs) are powerful black-box systems, making it difficult to discern whether their answers reflect stable internal beliefs or superficial pattern matching.
We identify \textbf{cross-contextual consistency} as an underutilized behavioral property of LLMs: a credible answer should remain stable when the same task is placed under topic-aligned, content-neutral contextual variation. Building on this intuition, we operationalize Cross-Contextual Consistency (C3) by comparing model generations under original and perturbed prompts. Across 26 models and six benchmarks spanning reasoning, factuality, and code generation, we find that answers with smaller cross-contextual shifts are more likely to be correct or factual. We demonstrate that C3 provides a complementary axis of evaluation and can serve as a benchmark usefulness diagnostic, identifying which portions of a benchmark remain informative even when aggregated scores are widely considered ``saturate".
\end{abstract}

\section{Introduction}

% \note{model credibility $\iff$ witness credibility in jury trials}

% \note{incorrect answers stay incorrect after perturbation}

Large language models (LLMs) achieve strong performance across a wide range of tasks, yet the internal basis of their responses remains poorly understood. 
In contrast to this strong performance, LLMs often fail to preserve logical consistency: For example, the \emph{reversal curse} \citep{berglund2024the} illustrates the failure to correctly associate facts that logically entail one another, while \emph{context hijacking} \citep{jiang2024llms} shows that appending semantically neutral context can trick LLMs into changing their answers. These examples show that despite their incredible performance on a variety of challenging tasks, LLMs can still fail at basic logical reasoning.
% To assess this, standard evaluations assess models by querying them directly, but such evaluations provide limited insight into whether the resulting answers are reliable or reflect stable internal beliefs. 
What's more, these failures are often unpredictable and manifest in surprising and unusual behaviours.
This motivates the need for principled approaches to directly assess model credibility and its relationship to performance and accuracy on downstream tasks.

% Despite their black-box nature, certain 
One popular approach is to exploit certain internal signals of large language models for evaluation. Token-level probabilities, for example, are commonly used in multiple-choice benchmarks to assess calibration~\citep{kapoor2024large, pmlr-v239-ren23a}. As powerful as these approaches can be in principle, these methods cannot be applied on closed-source models and, more critically, cannot handle free-form questions and open-ended responses. Other widely used methods such as self-report~\citep{lin2022teaching}, self-consistency~\citep{DBLP:conf/iclr/0002WSLCNCZ23}, and paraphrasing consistency~\citep{portillo-wightman-etal-2023-strength} are straightforward to implement, but each can suffer from mechanistic failures such as overconfidence~\citep{rathi2025humansoverrelyoverconfidentlanguage}, trapping on the same incorrect answer~\citep{chen-etal-2023-adaptation}, or changing of the prompt’s meaning~\citep{chataigner2025saywayauditingllms}, leading to inflated confidence estimates.

% \hilite{This paper}\note{which paper?} follows another line of work that adopts a more macroscopic, \hilite{behavior-level} approach, studying how LLM outputs vary under different prompting strategies and sampling techniques.

An alternative approach is to attempt to elicit the confidence of LLM responses \emph{indirectly}, without probing model parameters or asking for confidence measurements directly. This approach can be likened to placing a model under judicial scrutiny. As in legal proceedings, credibility is established not by a single answer but through \textbf{cross-examination}: deposition, repeated questioning, and extensive fact checking. Accordingly, a model’s response should be examined for consistency under alternative but semantically equivalent formulations of the same question. If a model truly \emph{affirms} its answer, it should be robust to cross-examination under logically and semantically equivalent contexts.
However, if a model’s outputs vary substantially across different framings, the credibility of the model is undermined.

\begin{figure*}[t]
    \centering
    \includegraphics[width=\textwidth]{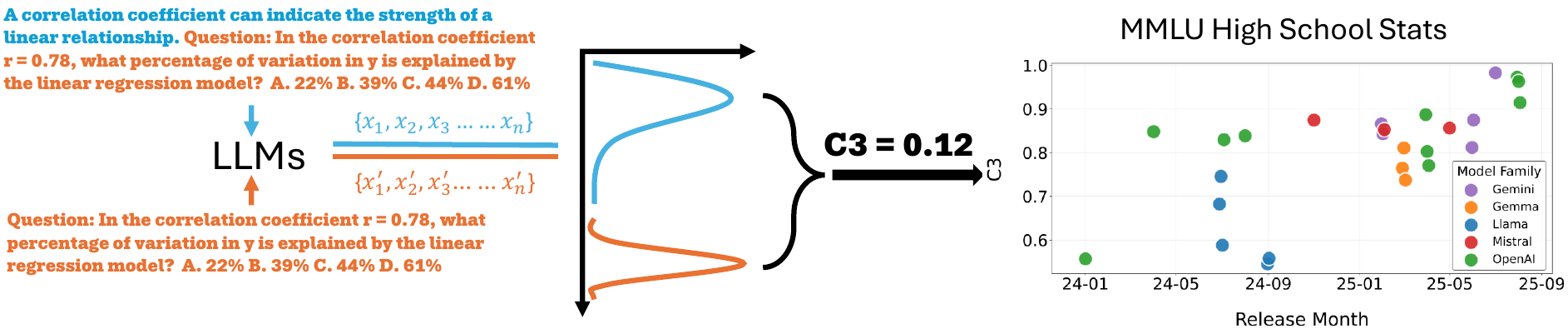}
    \caption{Overview of the proposed workflow. From left to right, we sample generations from the original prompt and from semantically neutral perturbed prompts, measure the resulting distributional shift, and compute C3. The right panel shows C3 for 26 models plotted by release month, revealing an increasing trend on the MMLU High School Statistics benchmark: newer models become progressively more credible.}
    \label{fig:workflow}
\end{figure*}

In this paper, we study a simple behavioral principle: an answer is more credible when it remains stable across contexts that change the surrounding wording or premise but not the task-relevant meaning. In an LLM, we call this property Cross-Contextual Consistency (C3), illustrated in Figure~\ref{fig:workflow}. 
By sampling a large number of prompt variations that preserve the original semantic content and then measuring the resulting distributional differences in model responses, we translate this behavioral principle into a quantitative metric for evaluating the credibility of a model. Rather than judging the correctness of individual answers, we evaluate the consistency of a model’s responses, which is a much simpler objective that 
relies less on external knowledge,
does not require access to model parameters, and
does not rely on self-reported measures.
Surprisingly, despite not being explicitly designed to test correctness---indeed, factuality and correctness are never explicitly evaluated by this metric---C3 proves to be a useful indicator of model correctness.

This metric, which is measured at the instance level, can be aggregated into a model-level robustness profile across different benchmarks. Evaluation across 16 models and six benchmarks shows that perturbation-based probing not only provides a practical means of quantifying how much a model’s generations shift under controlled, content-neutral prompt perturbations, but also reveals a consistent association between local stability and downstream reliability: Models (and individual task instances) that exhibit smaller distributional shifts under perturbation are more likely to produce correct answers, whereas larger shifts are frequently associated with errors. \hilite{\emph{This is despite the fact that accuracy is never explicitly measured by C3.}} Building on this observation, the results show that C3 provides a complementary evaluation axis that remains informative even when conventional benchmarks become less discriminative due to saturation and contamination.

Our main contributions are:

\begin{enumerate}[]
    \item \textbf{An underutilized behavioral property of LLMs.} We identify C3 as an underutilized behavioral property of LLMs' credibility of generations: when a model’s answer is well-supported, it should remain stable under cross-examination.

    \item\textbf{An operationalization of cross-contextual consistency.} We instantiate this idea through C3, a black-box evaluation protocol that compares model output distributions under original and perturbed contexts while adapting across multiple-choice, short-answer, long-form factuality, and code-generation tasks.
    
    \item \textbf{Empirical evidence for the evaluation utility of cross-contextual consistency.} Across six benchmarks and 16 main models, with an additional 10-model case study, we show that C3 aligns with correctness and factuality, remains robust across perturbation sources and comparison estimators, and provides a complementary diagnostic for identifying saturated, brittle, biased, and unlearned benchmark regions.
\end{enumerate}

% \section{A logical entailment view of inconsistency}
\section{Motivation: A tale from logical entailment}
\label{sec:logic}
When prompted with a question, what does it intuitively mean for an LLM to answer faithfully? One natural interpretation is that the model responds in accordance with the knowledge it possesses. Fortunately, this intuition can be formalized using the language of propositional logic. 

Let's write the internal parametric knowledge base of a model as \( \Gamma = \{\gamma_1, \ldots, \gamma_n\} \), consisting of \( n \) atomic binary formulae, from which \hilite{certain} statements produced by the model can be constructed via logical operators (e.g., conjunction, disjunction).  For example, we can have knowledge base with $\{\gamma_1 \text{=``Paris is in France"}, \gamma_2 \text{=``France is in Europe"} \}$ and statements like ``Paris is in an European country" ($\gamma_1 \land \gamma_2$) or ``France is a European country" ($\gamma_2$).
% Here, we assume that an LLM possesses an parametric knowledge base \( P = \{p_1, \ldots, p_n\} \), consisting of \( n \) atomic binary formulae, from which all statements produced by the model can be constructed.

An evaluation map $v$ assigns True or False values to atomic formulae, i.e. $v(\gamma_i)$, which induces truth values for statements built from atomic formulae.
A valuation $v$, under which a statement $f$ evaluates to true, is said to satisfy the statement, or to be a model of the statement. Let $M(f)$ be the set of models/worlds of $f$. Intuitively, $M(f)$ is the subset of all worlds under which statement $f$ is True.  Due to the stochastic nature of language models, each evaluation is associated with a probability $P(v)$ reflecting its likelihood of being true or false. In other words, rather than modeling an LLM as having a single coherent view, we model it as a mixture of views, each associated with a probability. 

Given two statements $f, g$, one can say that $f$ entails $g$ (i.e.,  $f \models g$) if $M(f) \subseteq M(g)$. That is, if $f$ entails $g$, then in every world where $f$ is True, $g$ must be True. For example, ``Paris is in an European country" entails ``France is a European country" with the knowledge base that contains ``Paris is in France" and ``France is in Europe" because $\gamma_1 \land \gamma_2 \models \gamma_2$. Interestingly, the two statements do not imply one another in isolation. The implication only holds once the internal knowledge base is taken into account.

When we do not have perfect entailment, the entailment probability can be computed as:
$
    P(f \models g) = 1 - \sum_{v \in M(f) \text{ and } v \not\in M(g)} P(v)
$
This is the complement of the probability mass assigned to worlds in which \(f\) is true and \(g\) is false. In the context of LLMs, we let $f$ denote a prompt and $g$ denote an response. We interpret $f \models g$ as meaning that the prompt $f$ entails the response $g$.

When evaluating LLMs, we seek to determine whether a given prompt-response pair is genuinely implied by the model’s internal world view, or, in stochastic terms, with what probability the model supports that implication. Under this framework, answering this question amounts to computing the aforementioned probability. 
However, two challenges remain: 
(1) How to sample different worlds $v$? 
(2) How to estimate different probabilities $P(v)$?

Our approach is motivated by considering different contexts as samples from different worlds. Each prompt corresponds to one world. Although the LLM’s output is stochastic, in practice the most probable answer typically dominates. This is because a single prompt may constrain induced behaviors. Therefore, we must perturb the prompts to induce diverse evaluation maps, while ensuring that the perturbed prompts remain within the semantic scope of \(M(f)\).

Still, directly estimating these underlying probabilities is ill-defined. Instead, we adopt a different approach. If for every world $v$ we have $f \models g$, then the entailment probability is $1$. Conversely, if there is substantial disagreement across worlds, the entailment probability is low. We therefore measure entailment by estimating the consistency of the model’s answers across different contexts, which can be interpreted as a notion of credibility. This is how we characterize (stochastic) entailment through cross-contextual consistency.

\section{Operationalizing Cross-Contextual Consistency}
\label{sec:c3}

Building on the motivation in \cref{sec:logic}, we operationalize \textbf{Cross-Contextual Consistency (C3)} as a behavioral probe for LLM credibility. The central intuition is simple: if a model's answer is well-supported, then its answer behavior should remain stable when the same task is placed in a different but answer-neutral context. Conversely, if the model's answer changes substantially under contextual variation that does not alter the task-relevant meaning, then the answer is more context-fragile. We therefore treat C3 as a cross-contextual comparison signal rather than as a single fixed metric: the goal is to compare model behavior under the original prompt and under controlled perturbed contexts.

\textbf{Controlled contextual perturbations.}
For each original task query $x \in X$, we sample a set of contextual perturbations
$\mathcal{E} = \{\epsilon_1, \dots, \epsilon_n\}$. Each perturbation $\epsilon \in \mathcal{E}$ is prefixed to the original query, producing a perturbed query $x' = [\epsilon; x]$. These perturbations are designed to change the surrounding context while preserving the task itself. In particular, we require perturbations to satisfy three properties: topic alignment, content neutrality, and non-trivial contextual variation. Topic alignment means that the perturbation remains within the broad domain or capability being tested. Content neutrality means that the perturbation does not reveal, support, contradict, or otherwise change the correct answer. Non-triviality means that the perturbation introduces meaningful contextual variation rather than merely restating the original question. We verify these three requirements separately in the appendix. Content neutrality is evaluated in Appendix~\ref{app:semantic neutrality}, where we examine whether the added context avoids introducing answer-relevant information or systematically shifting model performance. Topic alignment is examined in Appendix~\ref{app:Topic Alignment}, where we assess whether perturbations remain within the same broad domain or capability as the original query. Non-trivial contextual variation is verified in Appendix~\ref{app:diversity of noise}, where we check that accepted perturbations introduce diverse contextual variation rather than near-duplicate restatements.

Some examples of perturbed prompts are:
\begin{itemize}[leftmargin=*, itemsep=0pt, topsep=0pt, parsep=0pt]
    \item SVAMP: \emph{Many tourists visited the ancient castle during the weekend.} Rachel learned that 317 visitors came to Buckingham Palace that day. If there were 295 visitors the previous day, how many more visitors visited Buckingham Palace that day than on the previous day?
    \item SimpleQA: \emph{An artist might release an EP during any year of their career.} What EP did Rosalía release in 2019?
\end{itemize}

\textbf{Definition of C3.}
We formalize model generation as a stochastic process. Given an input $x$, an LLM induces a conditional distribution $P(Y \mid x)$ over possible outputs. This view follows the standard autoregressive perspective, where tokens are sampled sequentially based on the accumulated context and hidden state dynamics~\citep{NIPS2017_3f5ee243, Holtzman2020The, geshkovski2025mathematicalperspectivetransformers}. For an original query $x$, we sample a set of outputs
$\mathcal{Y} = \{y_1, y_2, \dots, y_n\} \sim P(Y \mid x)$.
For each contextual perturbation $\epsilon \in \mathcal{E}$, we construct a perturbed query
$x_\epsilon = [\epsilon; x]$. The perturbation set $\mathcal{E}$ induces a perturbed output distribution, $P^{\mathcal{E}}(Y \mid x) = \frac{1}{|\mathcal{E}|}\sum_{\epsilon \in \mathcal{E}} P(Y \mid x_\epsilon)$ from which we sample perturbed outputs $\mathcal{Y}^{\mathcal{E}} = \{y'_1, y'_2, \dots, y'_m\} \sim P^{\mathcal{E}}(Y \mid x)$.

Cross-Contextual Consistency (C3) measures how stable the model's answer behavior remains between the original and perturbed conditions. Formally, we define C3 as a normalized inverse distance between the original and perturbed output distributions: $\mathrm{C3}(x;\mathcal{E}) = 1 - \widetilde{D}\left(P(Y \mid x), P^{\mathcal{E}}(Y \mid x)\right)$,
where $\widetilde{D}(\cdot,\cdot)$ is a task-adaptive distance or disagreement function normalized to $[0,1]$. A higher C3 score indicates that the model's output distribution changes less under topic-aligned, content-neutral contextual variation, while a lower C3 score indicates greater context-fragility. This definition makes C3 a general cross-contextual comparison framework rather than a metric tied to a single distance.

\textbf{Distance metric: MMD.}
In our main implementation, we instantiate $\widetilde{D}$ using Maximum Mean Discrepancy (MMD)~\citep{JMLR:v13:gretton12a}. MMD provides a flexible non-parametric estimator for comparing empirical answer distributions, making it suitable for both fixed-format and open-ended generations. Given sampled outputs from the original condition $\mathcal{Y}$ and the perturbed condition $\mathcal{Y}^{\mathcal{E}}$, we compute an empirical MMD distance using task-adaptive feature maps and kernel choices. We then normalize the resulting distance into a consistency score in $[0,1]$, where larger values indicate smaller cross-contextual shift. Details on the empirical MMD estimator, task-adaptive feature maps, kernel choices, and normalization are provided in Appendix~\ref{app:Operationalization through MMD}.

Importantly, MMD provides one instantiation of C3, not the definition of C3 itself. C3 is defined by the comparison between model behavior under original and perturbed contexts. In Appendix~\ref{app:Ablation Study on MMD}, we replace MMD with a simpler cross-comparison distance and show that the same cross-contextual signal is largely preserved, supporting the view that C3 is driven by the original-versus-perturbed comparison rather than by a particular choice of distance.

\section{Experiments}
In this section, we describe our experimental setup, including data, methods, LLM models, and benchmarks.

\textbf{Data.}
We evaluate C3 on six widely used benchmarks spanning arithmetic reasoning, multiple-choice reasoning, commonsense inference, short-form factual QA, long-form factuality, and code generation. Specifically, we use SVAMP~\citep{patel-etal-2021-nlp}, MMLU High School Statistics~\citep{hendryckstest2021,hendrycks2021ethics}, CommonsenseQA~\citep{talmor-etal-2019-commonsenseqa}, SimpleQA Verified~\citep{haas2025simpleqaverifiedreliablefactuality}, FActScore~\citep{min-etal-2023-factscore}, and HumanEval~\citep{DBLP:journals/corr/abs-2107-03374}. Together, these benchmarks cover both fixed-format and open-ended generations, allowing us to evaluate C3 across diverse task formats and answer types. Detailed benchmark descriptions and prompts are provided in Appendix~\ref{app:benchmark_details_prompts}.

\textbf{Baselines.}
Since C3 is intended to serve as a proxy for the credibility of model generations, we compare it with related notions of confidence, consistency, and factuality. We include both vanilla black-box confidence estimators and a non-vanilla factuality checking tool. The vanilla baselines include \textbf{self-reported confidence}~\citep{lin2022teaching}, where the model outputs an explicit confidence score together with its answer; \textbf{self-consistency}~\citep{DBLP:conf/iclr/0002WSLCNCZ23}, which estimates confidence from agreement among repeated stochastic generations under the same prompt; and \textbf{paraphrasing consistency}~\citep{portillo-wightman-etal-2023-strength}, which measures whether answers remain stable under meaning-preserving prompt paraphrases. As a non-vanilla baseline, we compare against FActScore ~\citep{min-etal-2023-factscore}, which evaluates factual support for long-form generations using external evidence. For sampling-based baselines, agreement is computed with a generalized consistency score, using exact-match indicators for fixed-format tasks and semantic similarity for open-ended generations. Full implementation details are provided in Appendix~\ref{app:baseline_details}.

\textbf{Models.}
We evaluate C3 on 16 widely used LLMs spanning multiple families and scales. We also conduct case studies on 10 additional models; however, due to their characteristics, such as heavier reasoning processes or deprecated designs, they are often costly in time and computational resources. We therefore restrict these case studies to the MMLU High School Statistics benchmark. The full list of models is provided in Appendix~\ref{app:Models}.

\textbf{Evaluation metrics.}
We evaluate C3 and all baseline scores after normalizing each score to the range $[0,1]$. To measure calibration, we report the \textbf{Expected Calibration Error (ECE)}, where lower values indicate better calibration. For ranking-based evaluation, we report \textbf{AUROC}, as well as the area under the precision--recall curve for detecting correct outputs (\textbf{AUPRC-P}) and detecting incorrect outputs (\textbf{AUPRC-N}), with the latter computed using $1-s_i$. Details and implementation of these metrics are provided in Appendix~\ref{app:Evaluation Metrics}.

\begin{table*}[ht]
\centering

\scriptsize
\setlength{\tabcolsep}{2pt}
\renewcommand{\arraystretch}{1.3}

\begin{tabular*}{\linewidth}{@{\extracolsep{\fill}} l | cccc | cccc @{}}
\toprule 
& \multicolumn{4}{c|}{\textbf{CommonsenseQA}} & \multicolumn{4}{c}{\textbf{SVAMP}} \\
\textbf{Method}
& \makecell[c]{\scriptsize ECE $\downarrow$}
& \makecell[c]{\scriptsize AUROC $\uparrow$}
& \makecell[c]{\scriptsize AUPRC-P $\uparrow$}
& \makecell[c]{\scriptsize AUPRC-N $\uparrow$}
& \makecell[c]{\scriptsize ECE $\downarrow$}
& \makecell[c]{\scriptsize AUROC $\uparrow$}
& \makecell[c]{\scriptsize AUPRC-P $\uparrow$}
& \makecell[c]{\scriptsize AUPRC-N $\uparrow$} \\
\midrule
Self-consistency
& 0.224  &0.643    &0.756    &0.461
& 0.142  &0.861    &0.893    &0.740 \\
Paraphrasing
& 0.271  &0.634    &0.764    &0.484
& 0.157  &0.841    &0.873    &0.720 \\
Self-report
& 0.191  &0.542    &0.746    &0.332
& 0.240  &0.532    &0.763    &0.277 \\

C3 (Ours)
& \cellcolor{gray!20}0.189  & \cellcolor{gray!20} 0.722  & \cellcolor{gray!20}0.805 & \cellcolor{gray!20}0.541
& \cellcolor{gray!20}0.072 & \cellcolor{gray!20}0.917 & \cellcolor{gray!20}0.936 & \cellcolor{gray!20}0.818 \\
\midrule
\midrule
& \multicolumn{4}{c|}{\textbf{MMLU High School Statistics}} & \multicolumn{4}{c}{\textbf{FActScore}} \\
\textbf{Method}
& \makecell[c]{\scriptsize ECE $\downarrow$}
& \makecell[c]{\scriptsize AUROC $\uparrow$}
& \makecell[c]{\scriptsize AUPRC-P $\uparrow$}
& \makecell[c]{\scriptsize AUPRC-N $\uparrow$}
& \makecell[c]{\scriptsize ECE $\downarrow$}
& \makecell[c]{\scriptsize AUROC $\uparrow$}
& \makecell[c]{\scriptsize AUPRC-P $\uparrow$}
& \makecell[c]{\scriptsize AUPRC-N $\uparrow$} \\
\midrule
Self-consistency
& 0.382  &0.577    &0.511    &0.605
& 0.225  &0.858    &0.909    &\cellcolor{gray!20}0.731 \\
Paraphrasing
& 0.406  &0.547    &0.489    &0.591
& 0.249  &0.828    &0.887    &0.716 \\
Self-report
&  0.378  &0.496    &\cellcolor{gray!20}0.594    &0.411
& 0.330  &0.496    &0.706    &0.335 \\
C3 (Ours)
& \cellcolor{gray!20}0.338 &\cellcolor{gray!20}0.597    &0.530    &\cellcolor{gray!20}0.621
&\cellcolor{gray!20} 0.160  &\cellcolor{gray!20}0.873    &\cellcolor{gray!20}0.916    &0.701 \\
\midrule
\midrule
& \multicolumn{4}{c|}{\textbf{HumanEval}} & \multicolumn{4}{c}{\textbf{SimpleQA}} \\
\textbf{Method}
& \makecell[c]{\scriptsize ECE $\downarrow$}
& \makecell[c]{\scriptsize AUROC $\uparrow$}
& \makecell[c]{\scriptsize AUPRC-P $\uparrow$}
& \makecell[c]{\scriptsize AUPRC-N $\uparrow$}
& \makecell[c]{\scriptsize ECE $\downarrow$}
& \makecell[c]{\scriptsize AUROC $\uparrow$}
& \makecell[c]{\scriptsize AUPRC-P $\uparrow$}
& \makecell[c]{\scriptsize AUPRC-N $\uparrow$} \\
\midrule
Self-consistency
& 0.131  &0.812    &0.896    &0.487
& 0.393  &0.792    &0.368    &0.924 \\
Paraphrasing
& 0.165  &0.778    &0.862   &0.453
& 0.411  &0.773    &0.349    &0.906 \\
Self-report
& 0.287  &0.528    &0.807    &0.211
& 0.778 & 0.490 &  0.151  & 0.846 \\

C3 (Ours)
& \cellcolor{gray!20}0.110 & \cellcolor{gray!20}0.847 & \cellcolor{gray!20}0.943 & \cellcolor{gray!20}0.520
& \cellcolor{gray!20}0.166 & \cellcolor{gray!20}0.823 & \cellcolor{gray!20}0.389 & \cellcolor{gray!20}0.944 \\
\bottomrule
\end{tabular*}

\caption{Comparison of C3 with baselines across six benchmarks average over 16 models we tested. Shaded cells denote the best performance for each metric among the four methods. CommonsenseQA evaluates commonsense world knowledge; SVAMP evaluates mathematical reasoning; MMLU High School Statistics evaluates statistical knowledge; FactScore and SimpleQA evaluate factuality in long-form and short-form generation, respectively; and HumanEval evaluates code generation. Lower is better for ECE, while higher is better for AUROC, AUPRC-P, and AUPRC-N. The noise for perturbation in the table is sampled by GPT-4.1, and we also shown that C3 does not rely on advanced model and the results can be still reproducible by smaller models with very few costs as we shown in Appendix\ref{app:Ablation Study on Source of Noises}.}
\label{tab:main_results}
\end{table*}

\textbf{C3 elicitation.}
For both perturbed and unperturbed sampling, we collected 30 trials per instance across 16 standard models to assess the alignment of C3 against other baselines. The number 30 is supported by an empirical study shown in Appendix~\ref{app:number of trials}. In this study we use GPT-4.1 for noise sampling, however, an ablation study in Appendix~\ref{app:Ablation Study on Source of Noises} shows that perturbations from much smaller models are nearly as effective as those from larger ones, but with much lower computational overhead. The noise was prefixed to each prompt, separated by a single space. To determine answer equivalence, we use an LLM through the prompt detailed in Appendix~\ref{app:prompt for LLMs as Judges}. Given the high volume of evaluations required (tens of millions given the scale of our experiments), we performed offline inference using a Qwen3-8B~\citep{qwen3technicalreport} model temperature set to 0, supported by the VLLM framework~\citep{kwon2023efficient} on two NVIDIA H200 GPUs.

\vspace{-0.5em}
\section{Results}
We find that C3 provides substantially better signals than the baseline methods, and aligns especially well on challenging benchmarks such as SimpleQA, where other approaches often produce inflated assessments.

\subsection{C3 is calibrated with truthfulness}

\paragraph{Correctness.} The results in Table~\ref{tab:main_results} demonstrate that C3 consistently outperforms existing baselines in aligning model ``credibility'' with truthfulness.  In math tasks like SVAMP, C3 achieves an AUROC of 0.917, a noticeable jump over Self-Consistency, with AUROC of 0.861. This suggests that measuring the credibility of LLMs with perturbations is more effective for capturing the logical coherence of a reasoning chain than simple sampling strategies using identical prompts.

\textbf{Factuality.} Furthermore, the consistent gains on FActScore, CommonsenseQA, and SimpleQA suggest that C3 is effective for detecting factuality-related errors. In these settings, traditional consistency metrics can be poorly calibrated because models may remain highly consistent even when they are confidently relying on incorrect parametric memories or strong priors. By measuring the distributional shift in generations induced by content-neutral noise, C3 increases contrast between well-known facts, small shift, and poorly known facts, large shift, yielding a more informative signal of the model’s knowledge state.
\begin{wrapfigure}[23]{L}{0.5\linewidth}
    \vspace{-10pt}
    \centering
    \includegraphics[width=\linewidth]{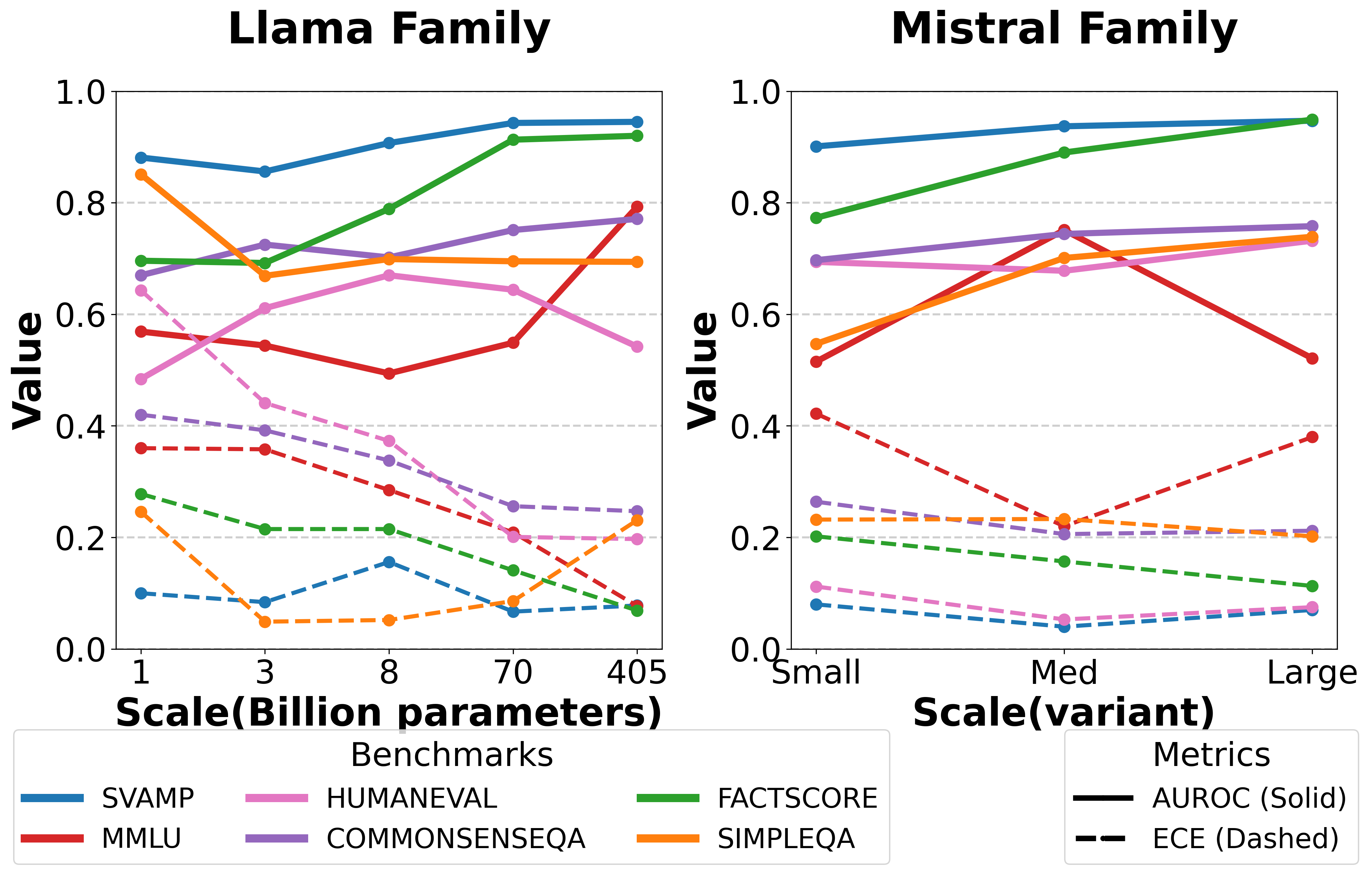}
    \caption{Scaling trends of Cross-Contextual Consistency (C3) calibration across benchmarks. Colors distinguish different benchmarks, while line styles represent metric types: solid lines denote AUROC (higher is better) and dashed lines denote ECE (lower is better). The results show that as model scale increases, the C3 becomes significantly better calibrated to correctness, evidenced by rising AUROC and declining ECE.}
    \label{fig:auroc_scaling_small_large_text}
    \vspace{-10pt}
\end{wrapfigure}

\textbf{Long form generation.}
Across the six benchmarks tested, HumanEval and FActScore are considered as ``long-form generation'' because they do not assume a pre-defined output format (such as multiple choice or keywords). C3 achieves nearly the best scores across four metrics in these two forms of generation: 0.110 ECE and 0.847 AUROC on HumanEval, and 0.160 ECE and 0.873 AUROC on FActScore. This is a scenario where many existing works fail. Our method scales cleanly to long-form generation, keeping nearly as simple as fixed-format tasks, which even allows us to evaluate code generation. While recent efforts like \citet{sharma2025assessingcorrectnessllmbasedcode} assess uncertainty in code via symbolic execution, our method achieves superior calibration without requiring external execution environments.

\textbf{Why they fail?}
Self-reported measures exhibit miscalibration: While they occasionally achieve seemingly reasonable ECE (e.g., 0.191 in CommonsenseQA), their discriminative power, measured by AUROC and AUPRC, remains near-random. This confirms that LLMs struggle to introspectively verbalize uncertainty, often overconfidently yielding near-100\% score for wrong answers~\citep{mielke-etal-2022-reducing, rathi2025humansoverrelyoverconfidentlanguage}. We also find that paraphrasing suffers from fundamental mechanistic flaws; since the paraphrasing is often performed by an LLM assistant, it can introduce semantic drift. For instance, high word overlap can mask cases where swapping arguments changes the underlying meaning~\citep{zhang-etal-2019-paws}, or the assistant may unintentionally alter the core intent of the prompt~\citep{chataigner2025saywayauditingllms}. Methods like self-consistency paraphrasing do not fail entirely aligning with truthfulness but are often trapped in the ``confidently wrong'' loop of LLM generation when the model sticks with an incorrect answer. As reflected in SimpleQA, which is a hard benchmark for many LLMs, self-consistency fails and is mis-calibrated with actual correctness on ECE, and paraphrasing makes results even worse. Measuring the distributional shift with C3, we successfully nudge the model to sample different answers when the knowledge is not grounded in LLMs' knowledge base, providing a much better signal in ECE and AUROC (0.166 and 0.823, respectively) in SimpleQA.

C3 becomes more informative as model capability increases. As shown in Figure~\ref{fig:auroc_scaling_small_large_text}, larger models within the Llama and Mistral families generally show stronger alignment between C3 and downstream correctness. This suggests that cross-contextual consistency is most useful once a model has enough capability for stable answer behavior to emerge; at that point, residual instability more clearly marks fragile answers rather than broad capability failure. However, scale does not eliminate cross-contextual fragility: even the strongest models remain imperfectly stable under answer-neutral contextual variation.

\subsection{A closer inspection on facutality}
\begin{wrapfigure}[26]{L}{0.5\linewidth}
    \vspace{-5pt}
    \centering
    \includegraphics[width=\linewidth]{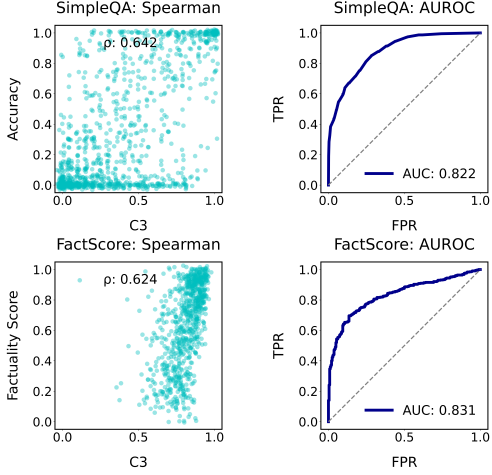}
    \caption{A detailed comparison of C3 alignment with factuality on FActScore and SimpleQA, showing Spearman rank correlation (left) and AUROC (right). C3 exhibits moderate to strong rank correlation with generation factuality. Minor AUROC differences from Table 1 are due to different aggregation strategies.}
    \label{fig:factuality}
    \vspace{-5pt}
\end{wrapfigure}

CommonsenseQA, FActScore, and SimpleQA all test a model's world knowledge and factuality. C3 shows a close alignment with the quality of generation across all three. While FActScore probes externally, scoring by decomposing long-form generation into atomic facts and verifying them against external sources, C3 probes internally. It uses variations of noise to perturb the LLM to see if it ``insists'' on an answer; the resulting distributional shift provides a significant signal regarding the quality of generation. As shown in Figure~\ref{fig:factuality}, when aggregating instance-level C3 and correctness, we observe a 0.624 Spearman rank correlation on FActScore and an 0.831 AUROC. We also include a comparison with SimpleQA, a benchmark where most models fail, making it a strong indicator for detecting overconfident metrics; C3 shows a 0.642 Spearman correlation and an 0.821 AUROC. Even without probing external information, C3 successfully aligns with FActScore. This suggests that the model’s internal state contains a latent representation of its own knowledge boundaries: when a model ``knows'' a fact, its output distribution is resilient to input noise, whereas hallucinated facts reside in low-probability regions that collapse or shift significantly under even minor perturbations.

\subsection{C3 for benchmark diagnosis}
\begin{wrapfigure}[37]{L}{0.5\linewidth}
    \vspace{-5pt}
    \centering
    \includegraphics[width=\linewidth]{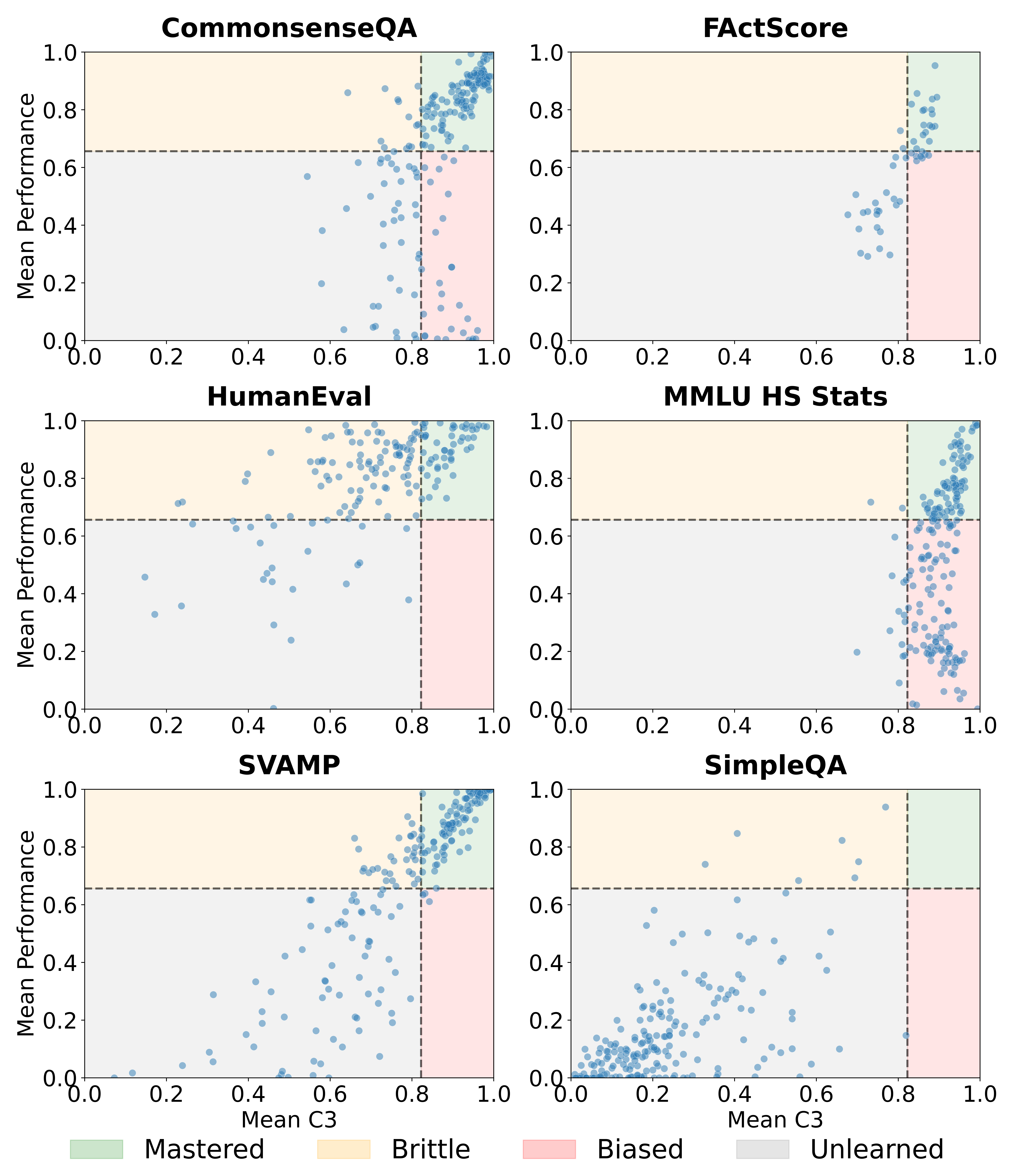}
    \caption{ For each benchmark instance, we compute the mean performance and mean C3 across the 16 models, then partition instances into four regions using the median performance and median C3 computed over instances across the six benchmarks. Green (“mastered”) indicates high performance with contextually consistent answers; yellow (“brittle”) indicates high performance but sensitive to perturbations; red (“biased”) indicates low performance yet perturbation-invariant (consistently wrong) answers; gray indicates low performance with inconsistency, suggesting knowledge that is not reliably learned.}
    \label{fig:Effect_analysis}
    \vspace{-5pt}
\end{wrapfigure}

By jointly inspecting instance-level C3 and instance-level benchmark performance, we can show cross-model typical behavior for each benchmark instance after averaging across the same 16 models. We find that C3 captures complementary information about model behavior beyond aggregate performance alone, as shown in Figure~\ref{fig:Effect_analysis}.

\textbf{Why do we need C3 as an additional axis?}
Relying exclusively on performance scores often mask the underlying mechanism of a model's knowledge. A key phenomenon in current LLM evaluation is that benchmarks may suffer from data contamination or overfitting, where models memorize specific prompt-answer pairs without acquiring the underlying reasoning. C3 provide additional information: ``Brittle'' (Yellow) region where models achieve high accuracy but fail to maintain consistency under perturbation. While high performance typically suggests capability, the low C3 in this region supports the alternative hypothesis: Success on these instances is driven by surface-level pattern matching rather than robust semantic understanding. This divergence serves as a proxy for detecting potential benchmark leakage to training processes of current LLMs.

C3 also helps differentiating mastered from systematic bias. The C3 axis further clarifies the status of the benchmark by distinguishing between ``solved'' and ``biased'' generation. Instances in the ``Mastered'' (Green) region represent tasks where models have converged on a stable solution. A high density of instances in this region supports the hypothesis of benchmark saturation, indicating that these specific questions no longer possess the discriminative power to distinguish between the capabilities of different models. Conversely, the ``Biased'' region highlights instances where models are not merely guessing, but are consistently trapped on incorrect answers. This supports the hypothesis that these benchmark instances trigger strong, incorrect priors or common misconceptions shared across models that possibly arise from model training processes. 

\textbf{Benchmark by Benchmark Comparison.}
Figure~\ref{fig:Effect_analysis} visualizes instance-level C3 against performance, revealing distinct patterns that characterize the status of each benchmark. CommonsenseQA and HumanEval both exhibit the signature of saturation, where a significant proportion of instances in the ``Mastered'' region suggests these tasks are well solved by modern models. However, in contrast to CommonsenseQA, HumanEval displays a heavy ``tail'' extending into the ``Brittle'' quadrant; this pattern implies that its high performance may be partially inflated by overfitting, where models succeed via surface pattern matching but fail under perturbation, lacking the true reasoning process of coding~\citep{riddell-etal-2024-quantifying}. SVAMP and FActScore show a balanced proportion of unlearned and mastered instances, while MMLU High School Stats shows a high number of mastered but also a significant cluster of biased instances; this suggests that while basics are understood, specific statistical concepts trigger consistent, systematic misconceptions. Finally, SimpleQA represents the true ``hard'' benchmark, dominated by the ``Unlearned'' region. The scarcity of mastered instances and the prevalence of stable poor performance indicate that this benchmark is not saturated, but rather captures specific knowledge gaps that remain out of reach for the current generation of models.

\section{Related work}\label{Related work}
A variety of methods to estimate model confidence have been proposed in the literature.
\textbf{White-box.} Methods rely on logits, hidden states, or parameters, which are infeasible for closed models; moreover, OpenAI reported degraded calibration after post-training, underscoring instability in probability-based confidence \citep{openai2024gpt4technicalreport,xie-etal-2024-calibrating,pmlr-v235-shen24c}. \
\textbf{Self-verbalized (black-box).} Such confidence can be miscalibrated and drift with prompting and post-training, and often shows over-confidence, even though some elicitation schemes help in specific settings \citep{lin2022teaching,xiong2024can,kumar-etal-2024-confidence,zhang-etal-2024-calibrating,heo2025do,openai2024gpt4technicalreport}. \
\textbf{Agreement-based (black-box).} Operating without input perturbation, plurality voting can conflate repetition bias with genuine certainty and underuse informative minority signals \citep{huang-etal-2024-mirror,DBLP:conf/iclr/0002WSLCNCZ23}. \
\textbf{Semantic paraphrase (black-box).} Paraphrase-based uncertainty presumes meaning preservation, yet small wording changes often shift semantics and behavior; entropy over paraphrases can therefore reflect semantic drift rather than true confidence \citep{zhang-etal-2019-paws,wahle-etal-2024-paraphrase,melamed-etal-2024-prompts,mizrahi-etal-2024-state}. Meanwhile paraphrasing with LLMs leads to biased collection of model performance, \citep{lunardi2025robustnessreliabilitybenchmarkbasedevaluation} shows that the  absolute accuracy
scores drop significantly when when paraphrased the quesiton. 
\textbf{Answer-calibration for long-form.} Treating confidence as probability of the correct answer’’ is ill-posed for multi-sentence, open-ended generations; automatic checks correlate unevenly with human judgment and preferences \citep{xu-etal-2023-critical,fabbri-etal-2021-summeval,chen-etal-2024-humans}.  \
\textbf{Other families (conformal, refusal).} These typically require labels, specialized scoring access, or finetuning—constraints that hinder post-hoc use in closed APIs \citep{quach2024conformal,pmlr-v235-mohri24a,zhang-etal-2024-r}.

\paragraph{Our approach}
We measure credibility via the distributional shift in generation induced by semantically equivalent, meaning-preserving prompt perturbations, using this perturbation gap as a reference-free proxy for confidence \citep{kuhn2023semantic,mizrahi-etal-2024-state}. This lens differs from existing prompt-sensitivity and robustness notions in several important ways. Some prior work quantifies sensitivity in probability space by tracking changes in log-likelihoods or likelihood ratios across prompt variants, which typically requires access to internal scoring signals and reflects probability drift rather than semantic drift of the generated content \citep{chatterjee-etal-2024-posix}. Others study robustness through format-induced variability by measuring sensitivity to spurious formatting features in prompt design, diagnosing evaluation volatility but not yielding an instance-level credibility signal for open-ended generations \citep{sclar2024quantifyinglanguagemodelssensitivity}. Related measures are also often defined for classification by analyzing instability of predicted label distributions under rephrasing, which does not transfer cleanly to free-form text where outputs must be compared semantically rather than as discrete labels \citep{errica-etal-2025-wrong}. Another line evaluates the reliability of knowledge probing methods by testing whether accept/reject decisions remain consistent under perturbations, emphasizing probe stability rather than the stability of the generated answer itself \citep{zhao-etal-2025-know}. In contrast, C3 directly measures invariance to meaning-preserving perturbations, is black-box and format-adaptive, and provides a comparable credibility diagnostic across heterogeneous tasks and open-ended generation formats.

\section{Conclusion}

This work identifies cross-contextual consistency as a behavioral signal for evaluating LLM credibility: when a model’s answer is well-supported, it should remain stable under topic-aligned, content-neutral contextual variation. We operationalize this idea through C3, a black-box framework that compares generation distributions under original and perturbed contexts. Across 26 models and six benchmarks, C3 consistently aligns with correctness and factuality across reasoning, factual recall, long-form generation, and code generation. Beyond instance-level evaluation, C3 also provides a complementary diagnostic for benchmark analysis, helping distinguish stable understanding from brittle or systematically biased performance. These results suggest that controlled contextual perturbation offers a practical way to reveal answer fragility that standard accuracy-based evaluation can miss.

\section{Limitations}

This work studies C3 across a broad but necessarily finite set of models, benchmarks, and experimental settings. Future work may extend the evaluation to additional tasks, languages, and application domains. Further exploration of alternative implementation choices may also help better understand how C3 can be adapted across different evaluation scenarios.

\section{Societal Impact}

This work aims to improve the evaluation of LLM reliability by providing a black-box signal for identifying context-fragile generations. Such tools may help practitioners better detect uncertain, brittle, or potentially hallucinated outputs before deploying LLMs in higher-stakes settings. At the same time, C3 should not be treated as a guarantee of truthfulness or safety; a model can be consistent and still wrong. In accordance with the NeurIPS Code of Ethics, we note that this method is intended as a diagnostic aid rather than a replacement for human oversight, domain expertise, or task-specific safety evaluation.
% \clearpage
% \newpage
\bibliographystyle{plainnat}
\bibliography{citation}

\newpage
\appendix
\onecolumn
\section{Discussion on Perturbation Noise source}\label{app:Discussion on Perturbation Noise source}

\subsection{LLMs as Sampler of Noise}\label{app:LLMs as Sampler of Noise}

We utilize LLMs as samplers to generate semantic perturbations. The noise generation process adheres to three critical principles to ensure the resulting samples are both challenging and informative:

\begin{itemize}
    \item \textbf{Topic Alignment:} Samples are generated to remain semantically aligned with the original input query to ensure the underlying task remains constant.
    \item \textbf{Content Neutrality:} The samples are designed to be content-neutral, as non-neutral content could introduce systematic bias or shifts in the model's performance.
    \item \textbf{Non-Trivial Contextual Variations:} The generated noise constitutes non-trivial contextual variations of the input, testing the model's robustness while maintaining the core meaning.
\end{itemize}

we employ the prompts detailed in Appendix~\ref{app:prompt for sampling}. By utilizing LLMs as samplers constrained by these principles, 
we collect a set of perturbations $\mathcal{E} = \{\epsilon_1, \dots, \epsilon_n\}$. 
To ensure the diversity of $\mathcal{E}$, we filter out candidate noise that is semantically redundant with existing entries (details are provided in Appendix~\ref{app:diversity of noise}).

\subsection{Diversity of Noise Sampling}\label{app:diversity of noise}
We utilize GPT-4.1 to perform noise sampling, aiming to maximize the diversity of noise per question. It is well known that repeated generations from LLMs using the same prompt often produce semantically similar or even identical outputs. To ensure variety, we maintain a clean pool of samples by comparing each newly generated data point $\epsilon^*$ against an existing set $\mathcal{E} = \{\epsilon_1, \epsilon_2, \dots, \epsilon_n\}$ in embedding space. A new sample $\epsilon^*$ is retained only if its pairwise semantic similarity with the existing elements in $\mathcal{N}$ remains below a threshold $k$.

We compute embeddings using \textit{all-mpnet-base-v2} \citep{song2020mpnet} and measure similarity via cosine similarity. However, the choice of similarity threshold $k$ is typically heuristic and task-specific. In this work, we propose to empirically determine an appropriate $k$ specific to GPT-4.1's generation behavior, enabling us to maintain a semantically diverse pool.

\begin{figure*}[ht] 
    \centering
    \includegraphics[width=\textwidth]{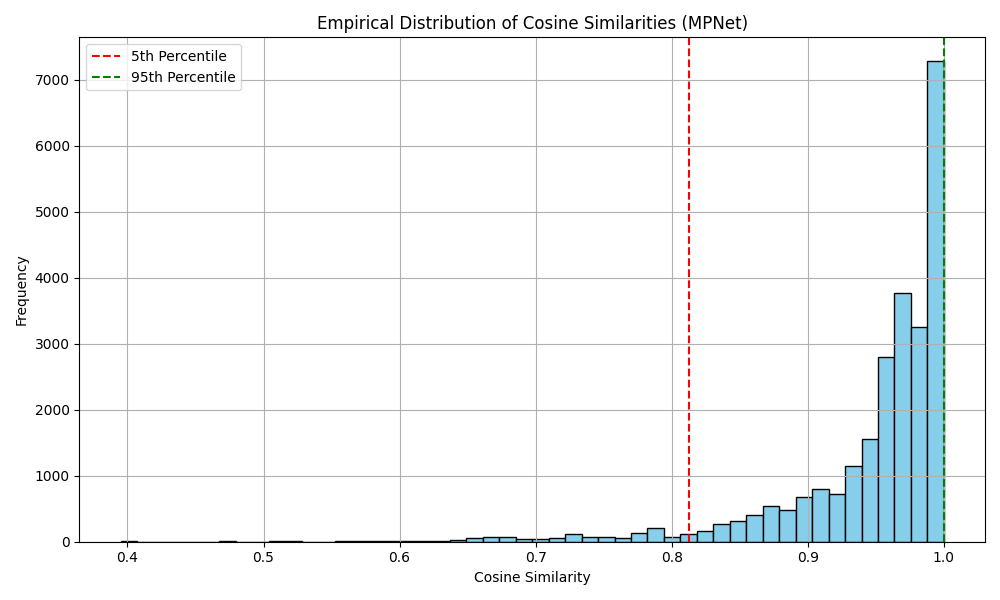}
    \caption{The distribution of pairwise cosine similarities among GPT-4.1-generated paraphrases. The red line indicates the 5th percentile ($k=0.8124$), which serves as our empirical threshold for filtering semantic redundancy.}
    \label{fig:rewrite distribution}
\end{figure*}

To do this, we sampled 30 premises generated by GPT-4.1 and used the same model to produce paraphrased versions using the prompt provided in Appendix \ref{app:prompt for paraphrasing}. For each original premise, we generated 50 paraphrases and embedded them using \textit{all-mpnet-base-v2}. Within each group of 50 rewrites, we computed all pairwise cosine similarities, resulting in $\binom{50}{2}$ similarity scores per group. Aggregating these across all groups yields the overall distribution of cosine similarities, representing GPT-4.1's implicit notion of semantic equivalence. The histogram of these results is shown in Figure \ref{fig:rewrite distribution}.

The threshold is determined at a significance level of $\alpha = 0.05$. This ensures that by retaining only those samples with a similarity below $k = 0.8124$, we have a statistical confidence that at most 5\% of the accepted samples are semantically redundant paraphrases.

\clearpage
\newpage

\subsection{Semantic Neutrality of Perturbations}
\label{app:semantic neutrality}

\begin{figure*}[ht] 
    \centering
    \includegraphics[width=\textwidth]{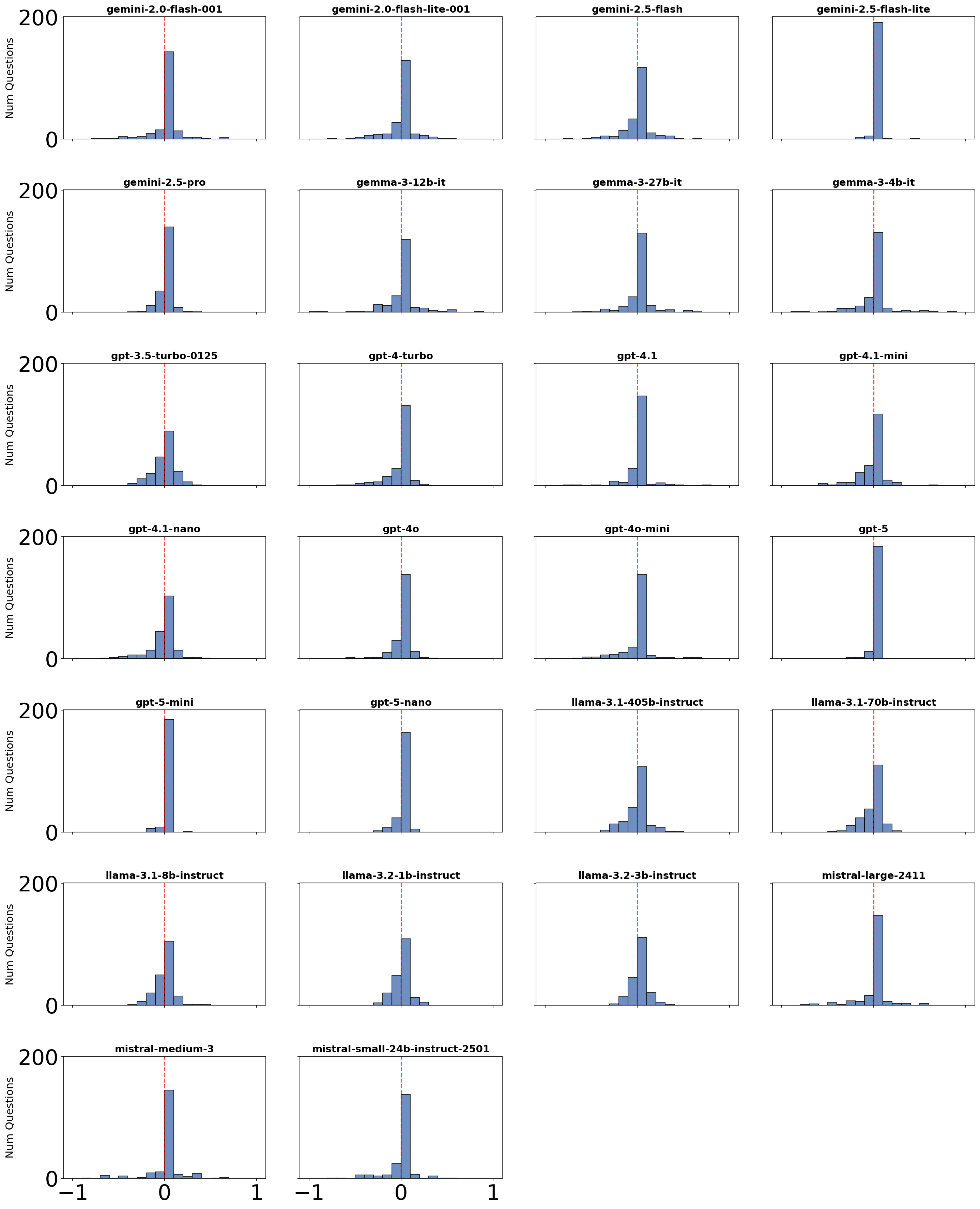}
    \caption{The difference of performance of each model before and after the perturbation noises are added. The red line denote the 0 meaning no difference in the performances.}
    \label{fig:neutrality}
\end{figure*}

Figure~\ref{fig:neutrality} evaluates whether our sampled perturbations introduce systematic answer-relevant bias. We conduct this check on MMLU High School Statistics across 26 models. For each model-question pair, we sample 30 generations from the original prompt and 30 generations from the perturbed prompt. We then compute:
\[
\Delta = \#\text{Correct}_{\text{perturbed}} - \#\text{Correct}_{\text{original}},
\]
where positive values indicate that the perturbation improves performance and negative values indicate that it hurts performance.

If the added context revealed information about the answer, contradicted the question, or otherwise biased the model toward or away from the correct option, we would expect the distribution of $\Delta$ to shift systematically above or below zero. In contrast, Figure~\ref{fig:neutrality} shows that the distribution is centered near zero across all 26 models, with no consistent positive or negative shift. This suggests that the perturbations do not systematically help or mislead the models on this benchmark.

This analysis does not prove that every individual perturbation is perfectly neutral, but it provides empirical evidence that the perturbation procedure does not introduce a systematic directional bias in model performance. We therefore treat these perturbations as approximately content-neutral for the purposes of measuring cross-contextual consistency.

\clearpage
\subsection{Topic Alignment}
\label{app:Topic Alignment}

We also evaluate whether the sampled perturbations remain topic-aligned with the benchmark from which they are generated. Topic alignment means that the perturbation should stay within the broad domain or capability being tested, without revealing, contradicting, or otherwise modifying the answer. This requirement separates our perturbations from arbitrary distractor text: the added context should create meaningful contextual variation, but should still be relevant to the type of task being evaluated.

To assess topic alignment, we conduct an automatic topic-classification check. For each benchmark, we used whole perturbations generated by our pipeline with GPT-4.1 and remove the corresponding original question and answer. We then ask an independent Qwen3-8B judge to classify each perturbation into one of the broad benchmark-level domains: arithmetic reasoning, statistical reasoning, commonsense reasoning, short-form factual recall, long-form factual or biographical generation, and code generation. The judge observes only the perturbation itself, not the original query, answer, or benchmark label. A perturbation is counted as topic-aligned if the judge assigns it to the same broad domain as the benchmark from which it was sampled.

Across the perturbations, approximately 96\% are classified into their intended benchmark domain. This suggests that the perturbations are not arbitrary out-of-domain distractors but instead remain aligned with the capability being evaluated. Together with the neutrality analysis in Appendix~\ref{app:semantic neutrality}, this supports the use of our sampled perturbations as topic-aligned, answer-neutral contextual variations for estimating cross-contextual consistency.
\clearpage
\newpage

\section{Empirical Choice of number of trials}\label{app:number of trials}
\begin{figure*}[ht] 
    \centering
    \includegraphics[width=\textwidth]{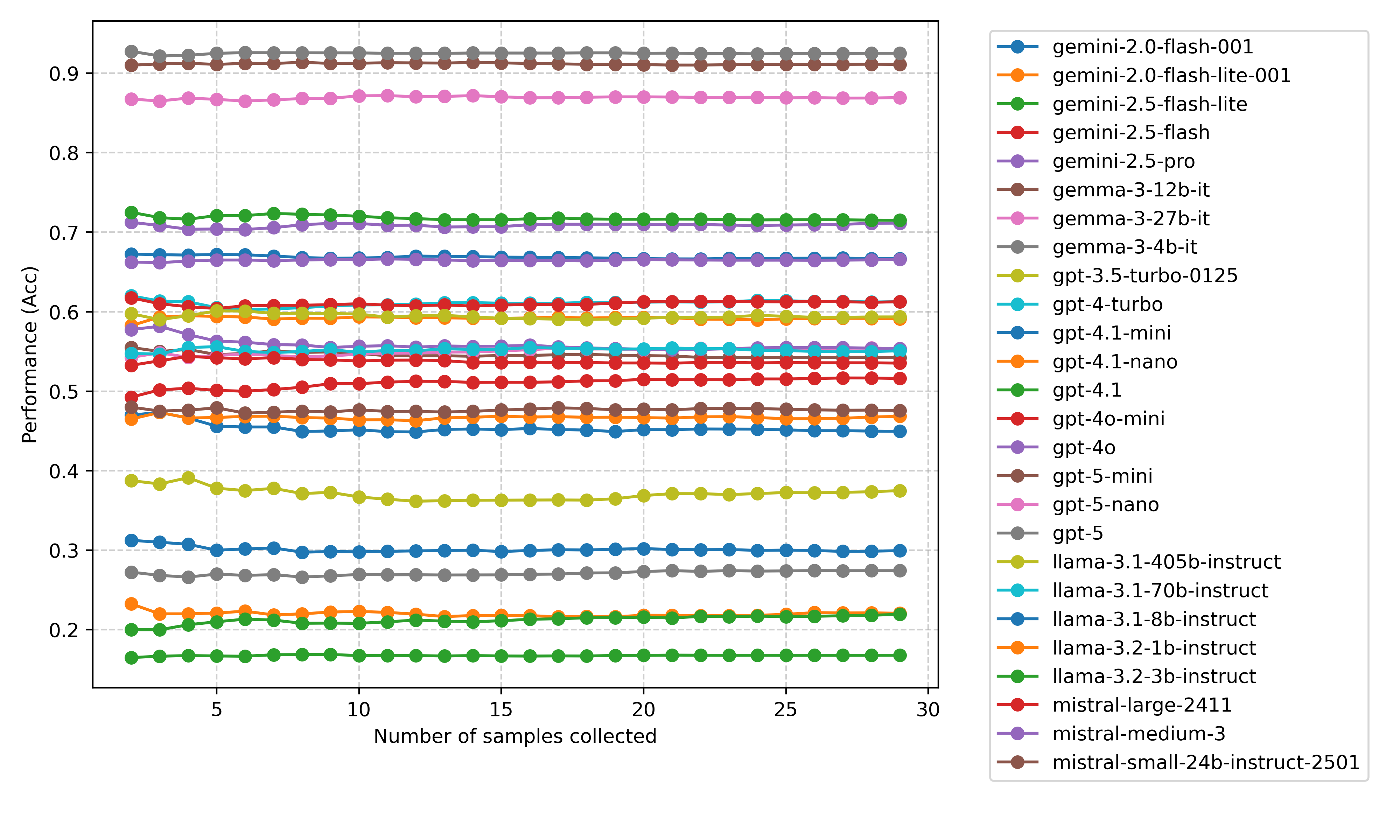}
    \caption{The performance of models across number of trials of samples we collected when perturbation is presented.}
    \label{fig:trial_convergence}
\end{figure*}

To choose an appropriate number of trials, we conducted a case study to determine how many samples per question are needed to reliably characterize model behavior before and after perturbation. As shown in Figure~\ref{fig:trial_convergence}, the estimated behavior stabilizes after about 20 samples. We therefore use 30 trials per question as a more conservative choice in this setting.
\clearpage
\newpage
\section{C3 Evaluation on Model-Level}\label{app:model level evaluation}
\begin{table}[H]
\centering
\caption{MMLU High School Stats Model Performance Metrics}
\label{tab:model-metrics}
\scriptsize % Reduces font size slightly to fit width comfortably
\begin{minipage}[t]{0.48\textwidth}
    \centering
    \begin{tabular}{lcccc}
        \toprule
        \textbf{Metric} & \textbf{ECE} & \textbf{AUROC} & \textbf{PR-P} & \textbf{PR-N} \\
        \midrule
        \multicolumn{5}{l}{\textbf{gemini-2.5-flash}} \\
        self-consistency & 0.359 & 0.506 & 0.527 & 0.495 \\
        Paraphrasing     & 0.394 & 0.456 & 0.452 & 0.472 \\
        C3              & 0.327 & 0.549 & 0.569 & 0.501 \\
        Self-report      & 0.271 & 0.492 & 0.741 & 0.237 \\
        \midrule
        \multicolumn{5}{l}{\textbf{gemini-2.5-flash-lite}} \\
        self-consistency & 0.795 & 0.503 & 0.166 & 0.835 \\
        Paraphrasing     & 0.808 & 0.450 & 0.150 & 0.789 \\
        C3            & 0.818 & 0.523 & 0.172 & 0.848 \\
        Self-report      & 0.374 & 0.519 & 0.579 & 0.442 \\
        \midrule
        \multicolumn{5}{l}{\textbf{gemma-3-12b-it}} \\
        self-consistency & 0.304 & 0.646 & 0.650 & 0.568 \\
        Paraphrasing     & 0.310 & 0.609 & 0.677 & 0.590 \\
        C3            & 0.241 & 0.679 & 0.689 & 0.592 \\
        Self-report      & 0.395 & 0.511 & 0.589 & 0.446 \\
        \midrule
        \multicolumn{5}{l}{\textbf{gemma-3-27b-it}} \\
        self-consistency & 0.326 & 0.647 & 0.663 & 0.543 \\
        Paraphrasing     & 0.329 & 0.608 & 0.660 & 0.555 \\
       C3             & 0.260 & 0.706 & 0.721 & 0.613 \\
        Self-report      & 0.423 & 0.458 & 0.549 & 0.420 \\
        \midrule
        \multicolumn{5}{l}{\textbf{gemma-3-4b-it}} \\
        self-consistency & 0.602 & 0.628 & 0.335 & 0.798 \\
        Paraphrasing     & 0.666 & 0.596 & 0.348 & 0.785 \\
       C3             & 0.504 & 0.595 & 0.324 & 0.776 \\
        Self-report      & 0.530 & 0.496 & 0.307 & 0.716 \\
        \midrule
        \multicolumn{5}{l}{\textbf{gpt-4.1}} \\
        self-consistency & 0.202 & 0.739 & 0.839 & 0.503 \\
        Paraphrasing     & 0.226 & 0.698 & 0.827 & 0.482 \\
       C3             & 0.180 & 0.753 & 0.856 & 0.528 \\
        Self-report      & 0.273 & 0.499 & 0.740 & 0.269 \\
        \midrule
        \multicolumn{5}{l}{\textbf{gpt-4.1-mini}} \\
        self-consistency & 0.416 & 0.536 & 0.493 & 0.561 \\
        Paraphrasing     & 0.418 & 0.506 & 0.494 & 0.506 \\
       C3             & 0.404 & 0.521 & 0.502 & 0.546 \\
        Self-report      & 0.233 & 0.592 & 0.879 & 0.232 \\
        \midrule
        \multicolumn{5}{l}{\textbf{gpt-4.1-nano}} \\
        self-consistency & 0.374 & 0.487 & 0.494 & 0.490 \\
        Paraphrasing     & 0.357 & 0.516 & 0.504 & 0.455 \\
       C3             & 0.358 & 0.493 & 0.510 & 0.502 \\
        Self-report      & 0.367 & 0.480 & 0.644 & 0.332 \\
        \bottomrule
    \end{tabular}
\end{minipage}
\hfill
\begin{minipage}[t]{0.48\textwidth}
    \centering
    \begin{tabular}{lcccc}
        \toprule
        \textbf{Metric} & \textbf{ECE} & \textbf{AUROC} & \textbf{PR-P} & \textbf{PR-N} \\
        \midrule
        \multicolumn{5}{l}{\textbf{llama-3.1-405b-instruct}} \\
        self-consistency & 0.122 & 0.784 & 0.849 & 0.618 \\
        Paraphrasing     & 0.164 & 0.757 & 0.828 & 0.633 \\
       C3             & 0.077 & 0.793 & 0.858 & 0.639 \\
        Self-report      & 0.300 & 0.508 & 0.743 & 0.274 \\
        \midrule
        \multicolumn{5}{l}{\textbf{llama-3.1-70b-instruct}} \\
        self-consistency & 0.211 & 0.598 & 0.657 & 0.500 \\
        Paraphrasing     & 0.209 & 0.536 & 0.617 & 0.523 \\
       C3             & 0.209 & 0.549 & 0.635 & 0.456 \\
        Self-report      & 0.276 & 0.520 & 0.747 & 0.271 \\
        \midrule
        \multicolumn{5}{l}{\textbf{llama-3.1-8b-instruct}} \\
        self-consistency & 0.366 & 0.490 & 0.308 & 0.686 \\
        Paraphrasing     & 0.395 & 0.480 & 0.260 & 0.625 \\
       C3             & 0.285 & 0.494 & 0.308 & 0.700 \\
        Self-report      & 0.489 & 0.491 & 0.373 & 0.632 \\
        \midrule
        \multicolumn{5}{l}{\textbf{llama-3.2-1b-instruct}} \\
        self-consistency & 0.492 & 0.538 & 0.220 & 0.832 \\
        Paraphrasing     & 0.540 & 0.517 & 0.180 & 0.773 \\
       C3             & 0.360 & 0.569 & 0.215 & 0.853 \\
        Self-report      & 0.576 & 0.517 & 0.163 & 0.862 \\
        \midrule
        \multicolumn{5}{l}{\textbf{llama-3.2-3b-instruct}} \\
        self-consistency & 0.470 & 0.477 & 0.192 & 0.814 \\
        Paraphrasing     & 0.476 & 0.480 & 0.198 & 0.759 \\
       C3             & 0.358 & 0.544 & 0.228 & 0.846 \\
        Self-report      & 0.540 & 0.477 & 0.286 & 0.696 \\
        \midrule
        \multicolumn{5}{l}{\textbf{mistral-large-2411}} \\
        self-consistency & 0.371 & 0.484 & 0.548 & 0.433 \\
        Paraphrasing     & 0.432 & 0.424 & 0.507 & 0.423 \\
       C3             & 0.380 & 0.521 & 0.573 & 0.463 \\
        Self-report      & 0.361 & 0.403 & 0.685 & 0.227 \\
        \midrule
        \multicolumn{5}{l}{\textbf{mistral-medium-3}} \\
        self-consistency & 0.286 & 0.629 & 0.731 & 0.445 \\
        Paraphrasing     & 0.317 & 0.600 & 0.654 & 0.493 \\
       C3             & 0.220 & 0.751 & 0.823 & 0.547 \\
        Self-report      & 0.274 & 0.499 & 0.772 & 0.242 \\
        \midrule
        \multicolumn{5}{l}{\textbf{mistral-small-24b}} \\
        self-consistency & 0.420 & 0.542 & 0.505 & 0.557 \\
        Paraphrasing     & 0.464 & 0.521 & 0.472 & 0.586 \\
       C3             & 0.422 & 0.515 & 0.491 & 0.526 \\
        Self-report      & 0.370 & 0.478 & 0.711 & 0.280 \\
        \bottomrule
    \end{tabular}
\end{minipage}
\end{table}

\clearpage
\begin{table}[H]
\centering
\caption{CommonsenseQA Model Performance Metrics}
\label{tab:cqa-metrics}
\scriptsize
\begin{minipage}[t]{0.48\textwidth}
    \centering
    \begin{tabular}{lcccc}
        \toprule
        \textbf{Metric} & \textbf{ECE} & \textbf{AUROC} & \textbf{PR-P} & \textbf{PR-N} \\
        \midrule
        \multicolumn{5}{l}{\textbf{gemini-2.5-flash}} \\
        self-consistency & 0.146 & 0.737 & 0.889 & 0.394 \\
        Paraphrasing     & 0.182 & 0.703 & 0.884 & 0.401 \\
       C3             & 0.150 & 0.741 & 0.903 & 0.381 \\
        Self-report      & 0.189 & 0.511 & 0.769 & 0.261 \\
        \midrule
        \multicolumn{5}{l}{\textbf{gemini-2.5-flash-lite}} \\
        self-consistency & 0.785 & 0.552 & 0.157 & 0.868 \\
        Paraphrasing     & 0.839 & 0.552 & 0.158 & 0.880 \\
       C3             & 0.808 & 0.584 & 0.168 & 0.892 \\
        Self-report      & 0.178 & 0.582 & 0.789 & 0.373 \\
        \midrule
        \multicolumn{5}{l}{\textbf{gemma-3-12b-it}} \\
        self-consistency & 0.228 & 0.571 & 0.755 & 0.358 \\
        Paraphrasing     & 0.261 & 0.569 & 0.764 & 0.396 \\
       C3             & 0.166 & 0.705 & 0.832 & 0.512 \\
        Self-report      & 0.219 & 0.493 & 0.723 & 0.309 \\
        \midrule
        \multicolumn{5}{l}{\textbf{gemma-3-27b-it}} \\
        self-consistency & 0.206 & 0.597 & 0.780 & 0.385 \\
        Paraphrasing     & 0.244 & 0.547 & 0.747 & 0.399 \\
       C3             & 0.154 & 0.731 & 0.854 & 0.539 \\
        Self-report      & 0.208 & 0.575 & 0.775 & 0.321 \\
        \midrule
        \multicolumn{5}{l}{\textbf{gemma-3-4b-it}} \\
        self-consistency & 0.334 & 0.558 & 0.654 & 0.444 \\
        Paraphrasing     & 0.369 & 0.509 & 0.646 & 0.440 \\
       C3             & 0.225 & 0.688 & 0.745 & 0.572 \\
        Self-report      & 0.338 & 0.476 & 0.588 & 0.409 \\
        \midrule
        \multicolumn{5}{l}{\textbf{gpt-4.1}} \\
        self-consistency & 0.146 & 0.650 & 0.852 & 0.419 \\
        Paraphrasing     & 0.181 & 0.637 & 0.875 & 0.440 \\
       C3             & 0.139 & 0.721 & 0.886 & 0.408 \\
        Self-report      & 0.110 & 0.561 & 0.841 & 0.226 \\
        \midrule
        \multicolumn{5}{l}{\textbf{gpt-4.1-mini}} \\
        self-consistency & 0.178 & 0.583 & 0.815 & 0.269 \\
        Paraphrasing     & 0.211 & 0.566 & 0.810 & 0.295 \\
       C3             & 0.156 & 0.691 & 0.863 & 0.462 \\
        Self-report      & 0.144 & 0.571 & 0.829 & 0.251 \\
        \midrule
        \multicolumn{5}{l}{\textbf{gpt-4.1-nano}} \\
        self-consistency & 0.228 & 0.662 & 0.796 & 0.401 \\
        Paraphrasing     & 0.262 & 0.704 & 0.852 & 0.453 \\
       C3             & 0.197 & 0.690 & 0.816 & 0.475 \\
        Self-report      & 0.158 & 0.551 & 0.728 & 0.412 \\
        \bottomrule
    \end{tabular}
\end{minipage}
\hfill
\begin{minipage}[t]{0.48\textwidth}
    \centering
    \begin{tabular}{lcccc}
        \toprule
        \textbf{Metric} & \textbf{ECE} & \textbf{AUROC} & \textbf{PR-P} & \textbf{PR-N} \\
        \midrule
        \multicolumn{5}{l}{\textbf{llama-3.1-405b-instruct}} \\
        self-consistency & 0.142 & 0.752 & 0.869 & 0.531 \\
        Paraphrasing     & 0.199 & 0.762 & 0.893 & 0.571 \\
       C3             & 0.117 & 0.791 & 0.894 & 0.576 \\
        Self-report      & 0.160 & 0.578 & 0.766 & 0.368 \\
        \midrule
        \multicolumn{5}{l}{\textbf{llama-3.1-70b-instruct}} \\
        self-consistency & 0.132 & 0.780 & 0.893 & 0.533 \\
        Paraphrasing     & 0.194 & 0.742 & 0.888 & 0.534 \\
       C3             & 0.115 & 0.813 & 0.924 & 0.525 \\
        Self-report      & 0.179 & 0.572 & 0.760 & 0.358 \\
        \midrule
        \multicolumn{5}{l}{\textbf{llama-3.1-8b-instruct}} \\
        self-consistency & 0.139 & 0.722 & 0.802 & 0.584 \\
        Paraphrasing     & 0.190 & 0.719 & 0.810 & 0.591 \\
       C3             & 0.110 & 0.745 & 0.851 & 0.573 \\
        Self-report      & 0.226 & 0.538 & 0.714 & 0.366 \\
        \midrule
        \multicolumn{5}{l}{\textbf{llama-3.2-1b-instruct}} \\
        self-consistency & 0.130 & 0.675 & 0.699 & 0.623 \\
        Paraphrasing     & 0.227 & 0.680 & 0.718 & 0.627 \\
       C3             & 0.093 & 0.684 & 0.718 & 0.607 \\
        Self-report      & 0.284 & 0.507 & 0.601 & 0.450 \\
        \midrule
        \multicolumn{5}{l}{\textbf{llama-3.2-3b-instruct}} \\
        self-consistency & 0.219 & 0.703 & 0.743 & 0.599 \\
        Paraphrasing     & 0.302 & 0.686 & 0.725 & 0.609 \\
       C3             & 0.128 & 0.738 & 0.797 & 0.637 \\
        Self-report      & 0.274 & 0.472 & 0.582 & 0.431 \\
        \midrule
        \multicolumn{5}{l}{\textbf{mistral-large-2411}} \\
        self-consistency & 0.189 & 0.577 & 0.799 & 0.329 \\
        Paraphrasing     & 0.225 & 0.618 & 0.849 & 0.380 \\
       C3             & 0.144 & 0.760 & 0.885 & 0.556 \\
        Self-report      & 0.137 & 0.576 & 0.843 & 0.241 \\
        \midrule
        \multicolumn{5}{l}{\textbf{mistral-medium-3}} \\
        self-consistency & 0.183 & 0.562 & 0.803 & 0.270 \\
        Paraphrasing     & 0.219 & 0.524 & 0.780 & 0.276 \\
       C3             & 0.146 & 0.742 & 0.881 & 0.484 \\
        Self-report      & 0.116 & 0.542 & 0.810 & 0.243 \\
        \midrule
        \multicolumn{5}{l}{\textbf{mistral-small-24b-instruct-2501}} \\
        self-consistency & 0.202 & 0.613 & 0.796 & 0.368 \\
        Paraphrasing     & 0.236 & 0.631 & 0.820 & 0.449 \\
       C3             & 0.172 & 0.722 & 0.861 & 0.460 \\
        Self-report      & 0.134 & 0.565 & 0.811 & 0.300 \\
        \bottomrule
    \end{tabular}
\end{minipage}
\end{table}
\clearpage
\begin{table}[H]
\centering
\caption{FactScore Model Performance Metrics}
\label{tab:factscore-metrics}
\scriptsize
\begin{minipage}[t]{0.48\textwidth}
    \centering
    \begin{tabular}{lcccc}
        \toprule
        \textbf{Metric} & \textbf{ECE} & \textbf{AUROC} & \textbf{PR-P} & \textbf{PR-N} \\
        \midrule
        \multicolumn{5}{l}{\textbf{gemini-2.5-flash}} \\
        self-consistency & 0.075 & 0.893 & 0.988 & 0.399 \\
        Paraphrasing     & 0.110 & 0.844 & 0.912 & 0.376 \\
       C3             & 0.050 & 0.924 & 0.992 & 0.464 \\
        Self-report      & 0.347 & 0.367 & 0.856 & 0.095 \\
        \midrule
        \multicolumn{5}{l}{\textbf{gemini-2.5-flash-lite}} \\
        self-consistency & 0.143 & 0.718 & 0.945 & 0.279 \\
        Paraphrasing     & 0.156 & 0.665 & 0.930 & 0.233 \\
       C3             & 0.069 & 0.741 & 0.957 & 0.247 \\
        Self-report      & 0.277 & 0.553 & 0.895 & 0.162 \\
        \midrule
        \multicolumn{5}{l}{\textbf{gemma-3-12b-it}} \\
        self-consistency & 0.343 & 0.915 & 0.920 & 0.926 \\
        Paraphrasing     & 0.349 & 0.877 & 0.947 & 0.948 \\
       C3             & 0.265 & 0.931 & 0.936 & 0.930 \\
        Self-report      & 0.411 & 0.443 & 0.522 & 0.416 \\
        \midrule
        \multicolumn{5}{l}{\textbf{gemma-3-27b-it}} \\
        self-consistency & 0.301 & 0.923 & 0.930 & 0.929 \\
        Paraphrasing     & 0.304 & 0.884 & 0.927 & 0.941 \\
       C3             & 0.222 & 0.973 & 0.981 & 0.971 \\
        Self-report      & 0.351 & 0.479 & 0.597 & 0.391 \\
        \midrule
        \multicolumn{5}{l}{\textbf{gemma-3-4b-it}} \\
        self-consistency & 0.411 & 0.935 & 0.936 & 0.948 \\
        Paraphrasing     & 0.475 & 0.903 & 0.949 & 0.935 \\
       C3             & 0.342 & 0.937 & 0.903 & 0.961 \\
        Self-report      & 0.361 & 0.560 & 0.468 & 0.699 \\
        \midrule
        \multicolumn{5}{l}{\textbf{gpt-4.1}} \\
        self-consistency & 0.080 & 1.000 & 1.000 & 1.000 \\
        Paraphrasing     & 0.103 & 0.959 & 0.988 & 0.979 \\
       C3             & 0.059 & 0.984 & 0.999 & 0.804 \\
        Self-report      & 0.289 & 0.332 & 0.892 & 0.068 \\
        \midrule
        \multicolumn{5}{l}{\textbf{gpt-4.1-mini}} \\
        self-consistency & 0.082 & 0.977 & 0.997 & 0.735 \\
        Paraphrasing     & 0.084 & 0.948 & 0.999 & 0.680 \\
       C3             & 0.064 & 0.989 & 0.999 & 0.915 \\
        Self-report      & 0.316 & 0.475 & 0.871 & 0.128 \\
        \midrule
        \multicolumn{5}{l}{\textbf{gpt-4.1-nano}} \\
        self-consistency & 0.136 & 0.788 & 0.969 & 0.633 \\
        Paraphrasing     & 0.119 & 0.817 & 0.979 & 0.597 \\
       C3             & 0.094 & 0.870 & 0.988 & 0.342 \\
        Self-report      & 0.356 & 0.492 & 0.936 & 0.093 \\
        \bottomrule
    \end{tabular}
\end{minipage}
\hfill
\begin{minipage}[t]{0.48\textwidth}
    \centering
    \begin{tabular}{lcccc}
        \toprule
        \textbf{Metric} & \textbf{ECE} & \textbf{AUROC} & \textbf{PR-P} & \textbf{PR-N} \\
        \midrule
        \multicolumn{5}{l}{\textbf{llama-3.1-405b-instruct}} \\
        self-consistency & 0.098 & 0.907 & 0.985 & 0.642 \\
        Paraphrasing     & 0.140 & 0.880 & 0.964 & 0.657 \\
       C3             & 0.069 & 0.920 & 0.988 & 0.540 \\
        Self-report      & 0.297 & 0.560 & 0.885 & 0.219 \\
        \midrule
        \multicolumn{5}{l}{\textbf{llama-3.1-70b-instruct}} \\
        self-consistency & 0.216 & 0.906 & 0.969 & 0.771 \\
        Paraphrasing     & 0.215 & 0.845 & 0.928 & 0.794 \\
       C3             & 0.141 & 0.913 & 0.971 & 0.780 \\
        Self-report      & 0.321 & 0.477 & 0.716 & 0.302 \\
        \midrule
        \multicolumn{5}{l}{\textbf{llama-3.1-8b-instruct}} \\
        self-consistency & 0.293 & 0.739 & 0.812 & 0.678 \\
        Paraphrasing     & 0.323 & 0.729 & 0.764 & 0.617 \\
       C3             & 0.215 & 0.789 & 0.817 & 0.737 \\
        Self-report      & 0.393 & 0.515 & 0.611 & 0.450 \\
        \midrule
        \multicolumn{5}{l}{\textbf{llama-3.2-1b-instruct}} \\
        self-consistency & 0.411 & 0.703 & 0.528 & 0.874 \\
        Paraphrasing     & 0.459 & 0.682 & 0.488 & 0.815 \\
       C3             & 0.278 & 0.696 & 0.580 & 0.837 \\
        Self-report      & 0.298 & 0.541 & 0.292 & 0.782 \\
        \midrule
        \multicolumn{5}{l}{\textbf{llama-3.2-3b-instruct}} \\
        self-consistency & 0.299 & 0.698 & 0.718 & 0.767 \\
        Paraphrasing     & 0.304 & 0.702 & 0.724 & 0.713 \\
       C3             & 0.215 & 0.692 & 0.702 & 0.719 \\
        Self-report      & 0.376 & 0.505 & 0.394 & 0.641 \\
        \midrule
        \multicolumn{5}{l}{\textbf{mistral-large-2411}} \\
        self-consistency & 0.206 & 0.929 & 0.985 & 0.804 \\
        Paraphrasing     & 0.266 & 0.868 & 0.944 & 0.793 \\
       C3             & 0.113 & 0.949 & 0.991 & 0.764 \\
        Self-report      & 0.305 & 0.543 & 0.834 & 0.273 \\
        \midrule
        \multicolumn{5}{l}{\textbf{mistral-medium-3}} \\
        self-consistency & 0.240 & 0.929 & 0.979 & 0.737 \\
        Paraphrasing     & 0.271 & 0.900 & 0.902 & 0.786 \\
       C3             & 0.157 & 0.890 & 0.968 & 0.599 \\
        Self-report      & 0.268 & 0.512 & 0.774 & 0.277 \\
        \midrule
        \multicolumn{5}{l}{\textbf{mistral-small-24b-instruct-2501}} \\
        self-consistency & 0.268 & 0.770 & 0.877 & 0.568 \\
        Paraphrasing     & 0.313 & 0.748 & 0.845 & 0.598 \\
       C3             & 0.202 & 0.773 & 0.879 & 0.613 \\
        Self-report      & 0.316 & 0.585 & 0.758 & 0.369 \\
        \bottomrule
    \end{tabular}
\end{minipage}
\end{table}

\clearpage
\begin{table}[H]
\centering
\caption{HumanEval Model Performance Metrics}
\label{tab:humaneval-metrics}
\scriptsize
\begin{minipage}[t]{0.48\textwidth}
    \centering
    \begin{tabular}{lcccc}
        \toprule
        \textbf{Metric} & \textbf{ECE} & \textbf{AUROC} & \textbf{PR-P} & \textbf{PR-N} \\
        \midrule
        \multicolumn{5}{l}{\textbf{gemini-2.5-flash}} \\
        self-consistency & 0.179 & 0.865 & 0.958 & 0.193 \\
        Paraphrasing     & 0.213 & 0.831 & 0.925 & 0.159 \\
       C3             & 0.165 & 0.911 & 0.998 & 0.237 \\
        Self-report      & 0.270 & 0.496 & 0.985 & 0.025 \\
        \midrule
        \multicolumn{5}{l}{\textbf{gemini-2.5-flash-lite}} \\
        self-consistency & 0.279 & 0.909 & 0.958 & 0.249 \\
        Paraphrasing     & 0.330 & 0.859 & 0.907 & 0.199 \\
       C3             & 0.237 & 0.941 & 0.999 & 0.282 \\
        Self-report      & 0.218 & 0.810 & 0.996 & 0.075 \\
        \midrule
        \multicolumn{5}{l}{\textbf{gemma-3-12b-it}} \\
        self-consistency & 0.087 & 0.782 & 0.916 & 0.493 \\
        Paraphrasing     & 0.119 & 0.750 & 0.883 & 0.461 \\
       C3             & 0.055 & 0.862 & 0.969 & 0.605 \\
        Self-report      & 0.311 & 0.433 & 0.810 & 0.152 \\
        \midrule
        \multicolumn{5}{l}{\textbf{gemma-3-27b-it}} \\
        self-consistency & 0.116 & 0.714 & 0.902 & 0.302 \\
        Paraphrasing     & 0.159 & 0.671 & 0.860 & 0.260 \\
       C3             & 0.069 & 0.768 & 0.954 & 0.360 \\
        Self-report      & 0.207 & 0.509 & 0.872 & 0.142 \\
        \midrule
        \multicolumn{5}{l}{\textbf{gemma-3-4b-it}} \\
        self-consistency & 0.093 & 0.740 & 0.823 & 0.635 \\
        Paraphrasing     & 0.134 & 0.699 & 0.782 & 0.594 \\
       C3             & 0.049 & 0.792 & 0.858 & 0.691 \\
        Self-report      & 0.294 & 0.535 & 0.721 & 0.315 \\
        \midrule
        \multicolumn{5}{l}{\textbf{gpt-4.1}} \\
        self-consistency & 0.110 & 0.857 & 0.964 & 0.236 \\
        Paraphrasing     & 0.140 & 0.826 & 0.934 & 0.205 \\
       C3             & 0.094 & 0.844 & 0.992 & 0.239 \\
        Self-report      & 0.272 & 0.618 & 0.966 & 0.079 \\
        \midrule
        \multicolumn{5}{l}{\textbf{gpt-4.1-mini}} \\
        self-consistency & 0.136 & 0.854 & 0.958 & 0.179 \\
        Paraphrasing     & 0.152 & 0.839 & 0.942 & 0.163 \\
       C3             & 0.113 & 0.883 & 0.995 & 0.190 \\
        Self-report      & 0.254 & 0.483 & 0.957 & 0.052 \\
        \midrule
        \multicolumn{5}{l}{\textbf{gpt-4.1-nano}} \\
        self-consistency & 0.100 & 0.812 & 0.951 & 0.547 \\
        Paraphrasing     & 0.129 & 0.783 & 0.922 & 0.518 \\
       C3             & 0.103 & 0.847 & 0.976 & 0.553 \\
        Self-report      & 0.273 & 0.529 & 0.924 & 0.100 \\
        \bottomrule
    \end{tabular}
\end{minipage}
\hfill
\begin{minipage}[t]{0.48\textwidth}
    \centering
    \begin{tabular}{lcccc}
        \toprule
        \textbf{Metric} & \textbf{ECE} & \textbf{AUROC} & \textbf{PR-P} & \textbf{PR-N} \\
        \midrule
        \multicolumn{5}{l}{\textbf{llama-3.1-405b-instruct}} \\
        self-consistency & 0.105 & 0.819 & 0.927 & 0.585 \\
        Paraphrasing     & 0.143 & 0.781 & 0.889 & 0.547 \\
       C3             & 0.104 & 0.806 & 0.944 & 0.560 \\
        Self-report      & 0.284 & 0.487 & 0.780 & 0.205 \\
        \midrule
        \multicolumn{5}{l}{\textbf{llama-3.1-70b-instruct}} \\
        self-consistency & 0.140 & 0.835 & 0.928 & 0.700 \\
        Paraphrasing     & 0.161 & 0.814 & 0.907 & 0.679 \\
       C3             & 0.113 & 0.852 & 0.956 & 0.656 \\
        Self-report      & 0.363 & 0.367 & 0.734 & 0.170 \\
        \midrule
        \multicolumn{5}{l}{\textbf{llama-3.1-8b-instruct}} \\
        self-consistency & 0.116 & 0.838 & 0.876 & 0.787 \\
        Paraphrasing     & 0.148 & 0.805 & 0.843 & 0.754 \\
       C3             & 0.093 & 0.875 & 0.918 & 0.841 \\
        Self-report      & 0.365 & 0.584 & 0.684 & 0.467 \\
        \midrule
        \multicolumn{5}{l}{\textbf{llama-3.2-1b-instruct}} \\
        self-consistency & 0.236 & 0.751 & 0.495 & 0.850 \\
        Paraphrasing     & 0.256 & 0.730 & 0.474 & 0.829 \\
       C3             & 0.130 & 0.851 & 0.754 & 0.917 \\
        Self-report      & 0.337 & 0.517 & 0.309 & 0.728 \\
        \midrule
        \multicolumn{5}{l}{\textbf{llama-3.2-3b-instruct}} \\
        self-consistency & 0.094 & 0.868 & 0.868 & 0.873 \\
        Paraphrasing     & 0.130 & 0.832 & 0.832 & 0.837 \\
       C3             & 0.089 & 0.871 & 0.874 & 0.874 \\
        Self-report      & 0.381 & 0.497 & 0.511 & 0.491 \\
        \midrule
        \multicolumn{5}{l}{\textbf{mistral-large-2411}} \\
        self-consistency & 0.100 & 0.772 & 0.937 & 0.452 \\
        Paraphrasing     & 0.138 & 0.733 & 0.898 & 0.413 \\
       C3             & 0.097 & 0.781 & 0.958 & 0.467 \\
        Self-report      & 0.245 & 0.583 & 0.906 & 0.154 \\
        \midrule
        \multicolumn{5}{l}{\textbf{mistral-medium-3}} \\
        self-consistency & 0.105 & 0.813 & 0.953 & 0.286 \\
        Paraphrasing     & 0.137 & 0.781 & 0.921 & 0.253 \\
       C3             & 0.121 & 0.862 & 0.981 & 0.401 \\
        Self-report      & 0.254 & 0.614 & 0.956 & 0.091 \\
        \midrule
        \multicolumn{5}{l}{\textbf{mistral-small-24b-instruct-2501}} \\
        self-consistency & 0.104 & 0.769 & 0.919 & 0.434 \\
        Paraphrasing     & 0.157 & 0.716 & 0.866 & 0.381 \\
       C3             & 0.123 & 0.802 & 0.958 & 0.445 \\
        Self-report      & 0.263 & 0.381 & 0.804 & 0.128 \\
        \bottomrule
    \end{tabular}
\end{minipage}
\end{table}

\clearpage
\begin{table}[H]
\centering
\caption{SimpleQA Model Performance Metrics}
\label{tab:simpleqa-metrics}
\scriptsize
\begin{minipage}[t]{0.48\textwidth}
    \centering
    \begin{tabular}{lcccc}
        \toprule
        \textbf{Metric} & \textbf{ECE} & \textbf{AUROC} & \textbf{PR-P} & \textbf{PR-N} \\
        \midrule
        \multicolumn{5}{l}{\textbf{gemini-2.5-flash}} \\
        self-consistency & 0.202 & 0.838 & 0.694 & 0.909 \\
        Paraphrasing     & 0.219 & 0.821 & 0.677 & 0.892 \\
       C3             & 0.114 & 0.827 & 0.633 & 0.924 \\
        Self-report      & 0.699 & 0.462 & 0.246 & 0.735 \\
        \midrule
        \multicolumn{5}{l}{\textbf{gemini-2.5-flash-lite}} \\
        self-consistency & 0.308 & 0.872 & 0.513 & 0.954 \\
        Paraphrasing     & 0.359 & 0.822 & 0.462 & 0.904 \\
       C3             & 0.129 & 0.835 & 0.356 & 0.974 \\
        Self-report      & 0.788 & 0.453 & 0.117 & 0.865 \\
        \midrule
        \multicolumn{5}{l}{\textbf{gemma-3-12b-it}} \\
        self-consistency & 0.487 & 0.781 & 0.112 & 0.970 \\
        Paraphrasing     & 0.502 & 0.765 & 0.096 & 0.955 \\
       C3             & 0.174 & 0.829 & 0.199 & 0.988 \\
        Self-report      & 0.898 & 0.448 & 0.058 & 0.928 \\
        \midrule
        \multicolumn{5}{l}{\textbf{gemma-3-27b-it}} \\
        self-consistency & 0.584 & 0.675 & 0.157 & 0.906 \\
        Paraphrasing     & 0.619 & 0.639 & 0.121 & 0.870 \\
       C3             & 0.244 & 0.754 & 0.231 & 0.961 \\
        Self-report      & 0.849 & 0.364 & 0.069 & 0.886 \\
        \midrule
        \multicolumn{5}{l}{\textbf{gemma-3-4b-it}} \\
        self-consistency & 0.540 & 0.785 & 0.057 & 0.983 \\
        Paraphrasing     & 0.570 & 0.754 & 0.026 & 0.952 \\
       C3             & 0.193 & 0.828 & 0.093 & 0.994 \\
        Self-report      & 0.952 & 0.669 & 0.040 & 0.986 \\
        \midrule
        \multicolumn{5}{l}{\textbf{gpt-4.1}} \\
        self-consistency & 0.239 & 0.762 & 0.659 & 0.807 \\
        Paraphrasing     & 0.249 & 0.752 & 0.648 & 0.796 \\
       C3             & 0.203 & 0.742 & 0.649 & 0.797 \\
        Self-report      & 0.559 & 0.456 & 0.355 & 0.609 \\
        \midrule
        \multicolumn{5}{l}{\textbf{gpt-4.1-mini}} \\
        self-consistency & 0.301 & 0.850 & 0.482 & 0.946 \\
        Paraphrasing     & 0.283 & 0.868 & 0.500 & 0.965 \\
       C3             & 0.136 & 0.864 & 0.558 & 0.968 \\
        Self-report      & 0.744 & 0.511 & 0.181 & 0.836 \\
        \midrule
        \multicolumn{5}{l}{\textbf{gpt-4.1-nano}} \\
        self-consistency & 0.233 & 0.897 & 0.415 & 0.999 \\
        Paraphrasing     & 0.241 & 0.889 & 0.406 & 0.991 \\
       C3             & 0.133 & 0.848 & 0.326 & 0.979 \\
        Self-report      & 0.812 & 0.451 & 0.090 & 0.912 \\
        \bottomrule
    \end{tabular}
\end{minipage}
\hfill
\begin{minipage}[t]{0.48\textwidth}
    \centering
    \begin{tabular}{lcccc}
        \toprule
        \textbf{Metric} & \textbf{ECE} & \textbf{AUROC} & \textbf{PR-P} & \textbf{PR-N} \\
        \midrule
        \multicolumn{5}{l}{\textbf{llama-3.1-405b-instruct}} \\
        self-consistency & 0.172 & 0.635 & 0.508 & 0.713 \\
        Paraphrasing     & 0.198 & 0.608 & 0.482 & 0.686 \\
       C3             & 0.231 & 0.674 & 0.517 & 0.745 \\
        Self-report      & 0.526 & 0.538 & 0.438 & 0.623 \\
        \midrule
        \multicolumn{5}{l}{\textbf{llama-3.1-70b-instruct}} \\
        self-consistency & 0.167 & 0.834 & 0.548 & 0.957 \\
        Paraphrasing     & 0.158 & 0.843 & 0.556 & 0.965 \\
       C3             & 0.086 & 0.890 & 0.600 & 0.972 \\
        Self-report      & 0.756 & 0.448 & 0.141 & 0.825 \\
        \midrule
        \multicolumn{5}{l}{\textbf{llama-3.1-8b-instruct}} \\
        self-consistency & 0.186 & 0.750 & 0.307 & 0.924 \\
        Paraphrasing     & 0.201 & 0.735 & 0.292 & 0.909 \\
       C3             & 0.052 & 0.806 & 0.286 & 0.974 \\
        Self-report      & 0.867 & 0.524 & 0.068 & 0.944 \\
        \midrule
        \multicolumn{5}{l}{\textbf{llama-3.2-1b-instruct}} \\
        self-consistency & 0.857 & 0.798 & 0.010 & 0.975 \\
        Paraphrasing     & 0.848 & 0.807 & 0.018 & 0.983 \\
       C3             & 0.246 & 0.942 & 0.125 & 0.999 \\
        Self-report      & 0.900 & 0.873 & 0.053 & 0.996 \\
        \midrule
        \multicolumn{5}{l}{\textbf{llama-3.2-3b-instruct}} \\
        self-consistency & 0.195 & 0.956 & 0.433 & 0.989 \\
        Paraphrasing     & 0.217 & 0.934 & 0.410 & 0.967 \\
       C3             & 0.049 & 0.947 & 0.300 & 0.998 \\
        Self-report      & 0.848 & 0.234 & 0.027 & 0.952 \\
        \midrule
        \multicolumn{5}{l}{\textbf{mistral-large-2411}} \\
        self-consistency & 0.628 & 0.761 & 0.422 & 0.911 \\
        Paraphrasing     & 0.655 & 0.734 & 0.394 & 0.883 \\
       C3             & 0.202 & 0.824 & 0.541 & 0.936 \\
        Self-report      & 0.720 & 0.464 & 0.214 & 0.753 \\
        \midrule
        \multicolumn{5}{l}{\textbf{mistral-medium-3}} \\
        self-consistency & 0.698 & 0.720 & 0.310 & 0.895 \\
        Paraphrasing     & 0.712 & 0.706 & 0.296 & 0.881 \\
       C3             & 0.233 & 0.810 & 0.516 & 0.944 \\
        Self-report      & 0.722 & 0.489 & 0.217 & 0.798 \\
        \midrule
        \multicolumn{5}{l}{\textbf{mistral-small-24b-instruct-2501}} \\
        self-consistency & 0.490 & 0.754 & 0.257 & 0.948 \\
        Paraphrasing     & 0.547 & 0.697 & 0.200 & 0.891 \\
       C3             & 0.232 & 0.754 & 0.299 & 0.949 \\
        Self-report      & 0.810 & 0.451 & 0.107 & 0.886 \\
        \bottomrule
    \end{tabular}
\end{minipage}
\end{table}
\clearpage
\begin{table}[H]
\centering
\caption{SVAMP Model Performance Metrics}
\label{tab:svamp-metrics}
\scriptsize
\begin{minipage}[t]{0.48\textwidth}
    \centering
    \begin{tabular}{lcccc}
        \toprule
        \textbf{Metric} & \textbf{ECE} & \textbf{AUROC} & \textbf{PR-P} & \textbf{PR-N} \\
        \midrule
        \multicolumn{5}{l}{\textbf{gemini-2.5-flash}} \\
        self-consistency & 0.047 & 0.934 & 0.986 & 0.766 \\
        Paraphrasing     & 0.062 & 0.914 & 0.966 & 0.746 \\
       C3             & 0.023 & 0.978 & 0.998 & 0.817 \\
        Self-report      & 0.106 & 0.523 & 0.894 & 0.150 \\
        \midrule
        \multicolumn{5}{l}{\textbf{gemini-2.5-flash-lite}} \\
        self-consistency & 0.121 & 0.942 & 0.976 & 0.783 \\
        Paraphrasing     & 0.136 & 0.922 & 0.956 & 0.763 \\
       C3             & 0.034 & 0.952 & 0.986 & 0.860 \\
        Self-report      & 0.216 & 0.500 & 0.784 & 0.216 \\
        \midrule
        \multicolumn{5}{l}{\textbf{gemma-3-12b-it}} \\
        self-consistency & 0.208 & 0.799 & 0.857 & 0.678 \\
        Paraphrasing     & 0.223 & 0.779 & 0.837 & 0.658 \\
       C3             & 0.075 & 0.913 & 0.946 & 0.834 \\
        Self-report      & 0.300 & 0.500 & 0.700 & 0.300 \\
        \midrule
        \multicolumn{5}{l}{\textbf{gemma-3-27b-it}} \\
        self-consistency & 0.178 & 0.833 & 0.909 & 0.649 \\
        Paraphrasing     & 0.193 & 0.813 & 0.889 & 0.629 \\
       C3             & 0.064 & 0.930 & 0.970 & 0.797 \\
        Self-report      & 0.240 & 0.500 & 0.760 & 0.240 \\
        \midrule
        \multicolumn{5}{l}{\textbf{gemma-3-4b-it}} \\
        self-consistency & 0.376 & 0.777 & 0.722 & 0.756 \\
        Paraphrasing     & 0.391 & 0.757 & 0.702 & 0.736 \\
       C3             & 0.117 & 0.846 & 0.835 & 0.805 \\
        Self-report      & 0.435 & 0.496 & 0.563 & 0.435 \\
        \midrule
        \multicolumn{5}{l}{\textbf{gpt-4.1}} \\
        self-consistency & 0.062 & 0.863 & 0.971 & 0.709 \\
        Paraphrasing     & 0.077 & 0.843 & 0.951 & 0.689 \\
       C3             & 0.026 & 0.927 & 0.985 & 0.787 \\
        Self-report      & 0.093 & 0.553 & 0.914 & 0.190 \\
        \midrule
        \multicolumn{5}{l}{\textbf{gpt-4.1-mini}} \\
        self-consistency & 0.078 & 0.872 & 0.964 & 0.724 \\
        Paraphrasing     & 0.093 & 0.852 & 0.944 & 0.704 \\
       C3             & 0.042 & 0.929 & 0.983 & 0.786 \\
        Self-report      & 0.101 & 0.605 & 0.909 & 0.150 \\
        \midrule
        \multicolumn{5}{l}{\textbf{gpt-4.1-nano}} \\
        self-consistency & 0.159 & 0.853 & 0.916 & 0.723 \\
        Paraphrasing     & 0.174 & 0.833 & 0.896 & 0.703 \\
       C3             & 0.090 & 0.878 & 0.943 & 0.688 \\
        Self-report      & 0.253 & 0.557 & 0.746 & 0.308 \\
        \bottomrule
    \end{tabular}
\end{minipage}
\hfill
\begin{minipage}[t]{0.48\textwidth}
    \centering
    \begin{tabular}{lcccc}
        \toprule
        \textbf{Metric} & \textbf{ECE} & \textbf{AUROC} & \textbf{PR-P} & \textbf{PR-N} \\
        \midrule
        \multicolumn{5}{l}{\textbf{llama-3.1-405b-instruct}} \\
        self-consistency & 0.032 & 0.953 & 0.993 & 0.711 \\
        Paraphrasing     & 0.047 & 0.933 & 0.973 & 0.691 \\
       C3             & 0.078 & 0.945 & 0.992 & 0.691 \\
        Self-report      & 0.128 & 0.547 & 0.881 & 0.186 \\
        \midrule
        \multicolumn{5}{l}{\textbf{llama-3.1-70b-instruct}} \\
        self-consistency & 0.062 & 0.927 & 0.980 & 0.693 \\
        Paraphrasing     & 0.077 & 0.907 & 0.960 & 0.673 \\
       C3             & 0.067 & 0.943 & 0.988 & 0.741 \\
        Self-report      & 0.165 & 0.512 & 0.838 & 0.175 \\
        \midrule
        \multicolumn{5}{l}{\textbf{llama-3.1-8b-instruct}} \\
        self-consistency & 0.053 & 0.904 & 0.902 & 0.901 \\
        Paraphrasing     & 0.068 & 0.884 & 0.882 & 0.881 \\
       C3             & 0.156 & 0.907 & 0.904 & 0.917 \\
        Self-report      & 0.362 & 0.538 & 0.639 & 0.419 \\
        \midrule
        \multicolumn{5}{l}{\textbf{llama-3.2-1b-instruct}} \\
        self-consistency & 0.141 & 0.856 & 0.751 & 0.908 \\
        Paraphrasing     & 0.156 & 0.836 & 0.731 & 0.888 \\
       C3             & 0.100 & 0.881 & 0.797 & 0.935 \\
        Self-report      & 0.458 & 0.613 & 0.568 & 0.578 \\
        \midrule
        \multicolumn{5}{l}{\textbf{llama-3.2-3b-instruct}} \\
        self-consistency & 0.151 & 0.844 & 0.715 & 0.905 \\
        Paraphrasing     & 0.166 & 0.824 & 0.695 & 0.885 \\
       C3             & 0.084 & 0.856 & 0.777 & 0.907 \\
        Self-report      & 0.418 & 0.509 & 0.573 & 0.430 \\
        \midrule
        \multicolumn{5}{l}{\textbf{mistral-large-2411}} \\
        self-consistency & 0.161 & 0.769 & 0.901 & 0.590 \\
        Paraphrasing     & 0.176 & 0.749 & 0.881 & 0.570 \\
       C3             & 0.070 & 0.947 & 0.984 & 0.790 \\
        Self-report      & 0.170 & 0.497 & 0.829 & 0.170 \\
        \midrule
        \multicolumn{5}{l}{\textbf{mistral-medium-3}} \\
        self-consistency & 0.225 & 0.796 & 0.869 & 0.633 \\
        Paraphrasing     & 0.240 & 0.776 & 0.849 & 0.613 \\
       C3             & 0.040 & 0.937 & 0.970 & 0.869 \\
        Self-report      & 0.166 & 0.556 & 0.846 & 0.256 \\
        \midrule
        \multicolumn{5}{l}{\textbf{mistral-small-24b-instruct-2501}} \\
        self-consistency & 0.222 & 0.851 & 0.880 & 0.703 \\
        Paraphrasing     & 0.237 & 0.831 & 0.860 & 0.683 \\
       C3             & 0.080 & 0.901 & 0.922 & 0.865 \\
        Self-report      & 0.230 & 0.500 & 0.770 & 0.230 \\
        \bottomrule
    \end{tabular}
\end{minipage}
\end{table}

\newpage
\section{Prompts}

\subsection{The Prompt for Sampling}\label{app:prompt for sampling}
\begin{PromptBox}{Prompt: LLMs as Noise Samplers}
You are given the following question: {question}.

Your task:

Identify the core topic/domain and the essential capability the question tests (e.g., arithmetic computation, set reasoning, causal inference).
Generate one short premise: a standalone sentence related to that same topic and capability that is neutral with respect to the question's main claim/condition (i.e., it neither entails nor contradicts it).

Requirements for the premise:
- Relevance: Stays within the same topic and capability as the question.
- Neutrality: Does not support, imply, or contradict the question's answer or key condition.
- Simplicity: 5--15 words, clear and concise.
- Randomness: Use different entities/names/numbers than those in the question; keep it generic; Obey Relevance and Neutrality, try your best to explore as much possibilities as possible.

Return ONLY a complete sentence.
\end{PromptBox}

\subsection{The Prompt for Sampling}\label{app:prompt for paraphrasing}
\begin{PromptBox}{Prompt: Paraphrasing}

Please rewrite the following statement to express the same meaning using different 
phrasing, while preserving its original logical content:\n\n{premise}

\end{PromptBox}
\subsection{The Prompt for Self-Confidence}\label{app:prompt for self confidence}
\begin{PromptBox}{Prompt: Self-Confidence}
Question: {question_prompt}

In addition, report your confidence in the correctness of this answer as a number between 0 and 100.
Note: The confidence indicates how likely you think your answer is true.

Respond strictly in the following format:
Answer: <Answer to the question>
Confidence: <Your confidence to the answer>
\end{PromptBox}

\subsection{The Prompt for Benchmarks}\label{app:prompt for benchmarks}
\begin{PromptBox}{Prompt: SVAMP}
Solve the following math word problem and provide only the final numeric answer.
Do not include steps, units, or explanation.

Problem:
Jessie weighed 92 kilograms. After she started to go jogging everyday she lost 56 kilograms in the first week and 99 kilograms in the second week. How much did she weigh after the first week of jogging?

 Final Answer:
\end{PromptBox}
\begin{PromptBox}{Prompt: MMLU High School Stats}
Question: The appraised values of houses in a city have a mean of $125,000 with a standard deviation of $23,000. Because of a new teachers' contract, the school district needs an extra 10% in funds compared to the previous year. To raise this additional money, the city instructs the assessment office to raise all appraised house values by $5,000. What will be the new standard deviation of the appraised values of houses in the city?
Choices:
A. $23,000 B. $25,300 C. $28,000 D. $30,300
Answer with a single letter (A-D).
Answer:
\end{PromptBox}

\begin{PromptBox}{Prompt: SimpleQA}
Question: How many television match officials were from England in the 2019 Rugby World Cup?
\end{PromptBox}

\begin{PromptBox}{Prompt: FActScore}
Tell me a bio of Kang Ji-hwan.
\end{PromptBox}
\begin{PromptBox}{Prompt: HumanEval}

from typing import List


def has_close_elements(numbers: List[float], threshold: float) -> bool:
    ``''`` Check if in given list of numbers, are any two numbers closer to each other than
    given threshold.
    >>> has_close_elements([1.0, 2.0, 3.0], 0.5)
    False
    >>> has_close_elements([1.0, 2.8, 3.0, 4.0, 5.0, 2.0], 0.3)
    True
    ''``''

\end{PromptBox}

\begin{PromptBox}{Prompt: CommonsenseQA}
Question: John was told to leave the cheese in the cellar for a few years.  Why is that?
Choices:
A. strong odor
B. age well
C. salad dressing
D. flavor
E. age to get better

Answer with a single letter (A-E).
\end{PromptBox}

\subsection{The Prompt for LLMs as Judges}\label{app:prompt for LLMs as Judges}
\begin{PromptBox}{Prompt: Judge Same Answer}
Task: Compare two code completions for the same problem.
Problem Description:
{problem}

Determine if the following two answers are essentially the same answer.
Ignore trivial differences like white space.

Answer A:
{ans_a}

Answer B:
{ans_b}

Answer strictly with '1' for Yes or '0' for No.
Answer:


\end{PromptBox}
\clearpage
\newpage
\section{Experiment Setup}
\subsection{Models}\label{app:Models}

We evaluate C3 on 16 widely used LLMs spanning multiple families and scales: OpenAI’s GPT-4.1 series (4.1, 4.1-mini, 4.1-nano)~\citep{openai2024gpt4technicalreport}; Google’s Gemma-3 Instruct models (4B, 12B, 27B)~\citep{gemmateam2025gemma3technicalreport} and Gemini models (Gemini-2.5-Flash, Gemini-2.5-Flash-Lite)~\citep{comanici2025gemini25pushingfrontier}; Meta’s Llama-3 Instruct models (1B, 3B, 8B, 70B, 405B)~\citep{grattafiori2024llama3herdmodels}; and Mistral models (Small, Medium, Large)~\citep{jiang2024mixtralexperts}.

We also conducted case studies on additional models; however, due to their characteristics: such as heavier reasoning processes or deprecated designs, they are often costly in time and computational resources. We therefore restrict these case studies to the MMLU High School Statistics benchmark. To study the temporal evolution of C3 in Figure~\ref{fig:workflow}, we additionally include earlier and newer frontier models, including GPT-3.5-Turbo~\citep{brown2020languagemodelsfewshotlearners}, GPT-4-Turbo, GPT-4o, and GPT-4o-mini~\citep{openai2024gpt4ocard}; GPT-5 (5, 5-mini, 5-nano)~\citep{singh2025openaigpt5card}; and additional Gemini releases (Gemini-2.5-Pro, Gemini-2.0-Flash, and Gemini-2.0-Flash-Lite)~\citep{geminiteam2025geminifamilyhighlycapable}. Across all experiments, we use temperature $T{=}1$ to probe typical stochastic generation behavior under standard decoding.
\subsection{Evaluation Metrics} \label{app:Evaluation Metrics}
We evaluate C3 and other baseline scores, all normalized to the range $[0, 1]$. A key aspect of these methods is to disclose the reliability of model generations: Their alignment with correctness or truthfulness. For each benchmark instance, we estimate an empirical model performance by aggregating outcomes over $n$ sampling trials (repeated samples of stochastic decoding) and taking the per-instance average. We then evaluate how well each score associated with instances aligns with this per-instance average performance using a combination of calibration and ranking metrics.

To measure calibration, we report the \textbf{Expected Calibration Error (ECE)}, computed by partitioning instances into $B$ bins according to their scores as $\mathrm{ECE}=\sum_{b=1}^{B}\frac{|I_b|}{N}\left|\mathrm{perf}(I_b)-\mathrm{score}(I_b)\right|$, where $N$ is the number of instances, $\mathrm{perf}(I_b)$ denotes the average performance of instances in bin $b$, and $\mathrm{score}(I_b)$ denotes the average predicted score in bin $b$; lower ECE values indicate better calibration. For ranking-based evaluation, we define binary labels by thresholding the per-instance average performance as $y_i=\mathbf{1}[\bar{p}_i \ge 0.5]$, and report \textbf{AUROC} by treating the score for instance $i$ as a ranking signal to separate instances with $y_i=1$ from those with $y_i=0$. In addition, we report precision-recall metrics that evaluate how well a score ranks instances by correctness, including the area under the precision--recall curve for detecting correct outputs (\textbf{AUPRC-P}) and for detecting incorrect outputs (\textbf{AUPRC-N}), with the latter using $1-s_i$.

\clearpage
\section{Ablation Study on Source of Noises}\label{app:Ablation Study on Source of Noises}

\begin{table}[h]

\centering

\begin{tabular*}{\linewidth}{@{\extracolsep{\fill}} l | cccc @{}}

\toprule
\multicolumn{5}{c}{\textbf{SimpleQA}} \\
\midrule
\textbf{Method}
& \makecell[c]{\scriptsize ECE $\downarrow$}
& \makecell[c]{\scriptsize AUROC $\uparrow$}
& \makecell[c]{\scriptsize AUPRC-P $\uparrow$}
& \makecell[c]{\scriptsize AUPRC-N $\uparrow$} \\
\midrule
Self-consistency & 0.393 & 0.792 & 0.368 & 0.924 \\
Paraphrasing     & 0.411 & 0.773 & 0.349 & 0.906 \\
Self-report      & 0.778 & 0.490 & 0.151 & 0.846 \\
C3 (GPT-4.1)        & \cellcolor{gray!20}0.166 & 0.823 & \cellcolor{gray!20}0.389 & \cellcolor{gray!20}0.944 \\
C3 (Qwen3-8B) & 0.235 & \cellcolor{gray!20}0.833 & 0.378 & 0.931 \\
C3 (Web Source) & 0.247 & 0.801 & 0.355 & 0.894 \\
\bottomrule
\end{tabular*}
\caption{Calibration results on SimpleQA comparing C3 under different perturbation sources. We compare perturbations generated by GPT-4.1, perturbations generated by the open model Qwen3-8B, and randomly sampled web-sourced noise. Results show that C3 remains competitive across perturbation sources, suggesting that its calibration signal is not solely dependent on frontier-model-generated perturbations.}
\label{tab:differen source of noise}
\end{table}

Although perturbations generated by GPT-4.1 are preferred in our main experiments because they consistently produce higher-quality generations and better satisfy the intended properties of diversity and content neutrality, C3 does not fundamentally depend on GPT-4.1 as the perturbation source. To test this, we compare GPT-4.1 perturbations with two alternative sources: (1) perturbations generated by the open-weight Qwen3-8B model, and (2) random web-sourced noise of approximately the same length sampled from RedPajama~\citep{together2023redpajama}, which contains internet-scraped text from a broad range of domains.

\textbf{Effectiveness of noise.}
As shown in Table~\ref{tab:differen source of noise}, C3 remains competitive when perturbations are generated from non-frontier sources. GPT-4.1 achieves the best ECE, AUPRC-P, and AUPRC-N, suggesting that higher-quality perturbations can improve calibration, especially in terms of probability calibration and precision-recall behavior. However, Qwen3-8B achieves the highest AUROC among the C3 variants and remains close to GPT-4.1 on AUPRC-P and AUPRC-N. Random web-sourced noise also preserves a meaningful calibration signal, outperforming or remaining competitive with the baseline methods on several metrics. These results suggest that the effectiveness of C3 is not solely an artifact of using a strong frontier model to generate perturbations. Instead, the core signal appears to come from measuring whether model outputs remain stable under semantically neutral contextual variation.

\textbf{Computational efficiency.}
It has to be admitted that using advanced model like GPT-4.1 can be costly and using smaller open-weight models substantially reduces the cost of perturbation generation. In our Qwen3-8B setting, we generated perturbations for 200 SimpleQA questions on a single GPU NVIDIA A100. We generated 30 samples per question, this corresponds to 6,000 accepted noise samples in 32.36 minutes after we applied filtering for diversity. This is the exact setting in the main experiment. The generation pipeline achieved 3.09 completed accepted samples per second. These results suggest that perturbation generation with smaller and open sourced model is practically feasible and can substantially reduce dependence on expensive frontier-model APIs.

Randomly sampled in-the-wild corpus noise provides an even cheaper alternative. Unlike model-generated perturbations, this source does not require inference or prompt-specific generation: once a corpus such as RedPajama~\citep{together2023redpajama} is available, noise snippets of the desired length can be sampled almost instantly. Because these snippets are sampled independently of the original question, they are unlikely to contain answer-specific information, which gives them a degree of semantic neutrality. However, this neutrality comes at the cost of weaker topic alignment: unlike GPT-4.1 or Qwen3-8B perturbations, in-the-wild snippets are not explicitly generated to match the benchmark domain or capability being tested.

As shown in Table~\ref{tab:differen source of noise}, corpus-based noise still preserves a useful C3 signal, but it is generally less effective than model-generated perturbations. This pattern suggests that topic alignment improves perturbation quality and strengthens the resulting C3 signal, while not being strictly necessary for C3 to remain informative. Web-sourced perturbations can therefore be viewed as a practical low-cost approximation when generation cost is a concern, rather than as a replacement for topic-aligned model-generated perturbations.

\clearpage
\section{Ablation Study on MMD}\label{app:Ablation Study on MMD}
\label{sec:mmd_ablation}

\begin{table}[h]
\centering
\caption{Calibration results on SimpleQA comparing the original MMD C3 score with a simpler cross-comparison methods. The cross-comparison method replaces the MMD distance with the average pairwise similarity between generations from the original and perturbed prompts. The noises are still sampled from GPT-4.1.}
\label{tab:mmd_ablation}
\begin{tabular*}{\linewidth}{@{\extracolsep{\fill}} l | cccc @{}}
\toprule
\multicolumn{5}{c}{\textbf{SimpleQA}} \\
\midrule
\textbf{Method}
& \makecell[c]{\scriptsize ECE $\downarrow$}
& \makecell[c]{\scriptsize AUROC $\uparrow$}
& \makecell[c]{\scriptsize AUPRC-P $\uparrow$}
& \makecell[c]{\scriptsize AUPRC-N $\uparrow$} \\
\midrule
Self-consistency       & 0.393 & 0.792 & 0.368 & 0.924 \\
Paraphrasing           & 0.411 & 0.773 & 0.349 & 0.906 \\
Self-report            & 0.778 & 0.490 & 0.151 & 0.846 \\
C3 (MMD)               & 0.166 & \cellcolor{gray!20}0.823 & 0.389 & \cellcolor{gray!20}0.944 \\
C3 (Cross Comparison)  & \cellcolor{gray!20}0.155 & 0.813 & \cellcolor{gray!20}0.402 & 0.931 \\
\bottomrule
\end{tabular*}
\end{table}

To test whether the effectiveness of C3 depends specifically on the MMD formulation, we replace the original MMD  distributional distance with a simpler cross-comparison score. Given generations from the original prompt $X=\{x_1,\ldots,x_n\}$ and generations from the perturbed prompt $Y=\{y_1,\ldots,y_m\}$, the cross-comparison variant directly measures the average pairwise agreement between the two sets:\[S_{\mathrm{cross}}(X,Y) = \frac{1}{nm}\sum_{i=1}^{n}\sum_{j=1}^{m} k(x_i,y_j),\]where $k(x_i,y_j)$ is an indicator function for answer equivalence:\[k(x_i,y_j)=\mathbf{1}[x_i \equiv y_j].\]
Here, $x_i \equiv y_j$ means that the two generations give the same answer. For fixed-format tasks such as SimpleQA, this can be implemented by exact answer matching or by an equivalence judge when surface forms differ but the answer is semantically the same. As shown in Table~\ref{tab:mmd_ablation}, the cross-comparison remains competitive with the original MMD C3 score. It slightly improves ECE and AUPRC-P on SimpleQA, while MMD achieves higher AUROC and AUPRC-N. More importantly, both C3 variants outperform self-consistency, paraphrasing consistency, and self-report on most calibration and ranking metrics. This suggests that the C3 signal is not merely an artifact of the specific MMD distance. Instead, the useful signal appears to come from the broader perturbation comparison: when a model's generations remain equivalent across original and semantically perturbed contexts, its answers are more likely to be reliable; when the cross-context generations diverge, the answer is more likely to be fragile or incorrect. MMD remains our main choice because it provides a principled distributional distance that accounts for both within-set and cross-set similarities, but this ablation shows that a simpler indicator cross-comparison variant can preserve much of the same credibility signal.

\clearpage
\section{Operationalization through MMD}\label{app:Operationalization through MMD}
To quantify the distance between the generative distributions $P(Y|x)$ and $P(Y|x')$, we use Maximum Mean Discrepancy (MMD)~\citep{JMLR:v13:gretton12a}, a non-parametric kernel-based statistic for comparing two empirical distributions. Given finite sample sets $\mathcal{Y}$ and $\mathcal{Y}'$ of size $n$ and $m$, we compute the unbiased empirical estimate:
\begin{equation}
\scriptsize
\begin{aligned}
    \widehat{\mathrm{MMD}}^2(\mathcal{Y}, \mathcal{Y}') 
    &= \frac{1}{n(n-1)} \sum_{i \neq j}^n 
    k(\phi(y_i), \phi(y_j)) 
    + \frac{1}{m(m-1)} \sum_{i \neq j}^m 
    k(\phi(y'_i), \phi(y'_j)) \\
    &\quad - \frac{2}{nm} \sum_{i=1}^n \sum_{j=1}^m 
    k(\phi(y_i), \phi(y'_j)).
\end{aligned}
\end{equation}
Here, $\phi$ maps model generations into a task-appropriate representation, and $k$ is a kernel function that compares represented outputs. This estimator measures how much the empirical answer distribution under the original prompt differs from the answer distribution under the perturbed prompt. In the main experiments, we normalize this distance into a C3 score in $[0,1]$, where larger values indicate smaller cross-contextual shift and therefore greater answer stability.

\textbf{Task-adaptive feature maps and kernel selection}
The flexibility of choices of feature mapping function $\phi(\cdot)$ and kernel function $k(\cdot, \cdot)$ provide the flexibility of assessing generation of various types. The realization of $\phi$ and $k$ are adapted to the specific format of the model's output $y$ of the underlying task. For tasks with fixed output formats (e.g,. keywords, numbers, or multiple-choices), we could choose $\phi$ to be a mapping to categories, resulting in an indicator kernel $k(y, y') = \mathbb{I}(y = y')$. This setting enables the C3 to function as an distance between empirical probability mass functions, measuring categorical inconsistencies. Conversely, for open-ended generation (e.g,. coding, summarization, and essay writing), $\phi$ could be a embedding process from an embedding model that maps the generation to high dimensional spaces $h \in \mathbb{R}^D$. In these high-dimensional space, we employ a semantic kernel (typically a linear dot-product to measure the cos similarities). Even more flexibly, LLMs as Judges frameworks can be used directly to compare the if the two answers are the same or not regardless of the output formats.

% \note{this reads too vague: what do you actually do? OK to discuss generalization, possibilities, etc as well}

\textbf{Normalization and range of C3}
To ensure C3 is an interpretable proxy for credibility, we transform the raw MMD distance into a normalized range of $[0, 1]$. In our framework, we assume a characteristic kernel $k$ that is bounded and normalized, satisfying $k(y, y) = 1$ and $0 \le k(y, y') \le 1$ (e.g., an indicator kernel or cosine similarity). Under these conditions, the squared MMD admits the theoretical upper bound $MMD^2(\mathcal{Y}, \mathcal{Y}') \leq 2$, which is attained when the two generative distributions are maximally separated (i.e., their cross-similarity approaches zero; equivalently, in fixed-format settings, outputs from the two sets completely mismatch). We define the final C3 by scaling this distance:
$
C3(x, x') = 1 - \frac{1}{2}\widehat{MMD}^2(\mathcal{Y}, \mathcal{Y}')
$
.A value of $C3 \approx 1$ indicates high credibility, where the model's output distribution remains invariant to semantic perturbations, suggesting consistent internal reasoning patterns and parametric memories. Conversely, $C3 \approx 0$ indicates low credibility, signaling that semantic variation in the input has caused a complete shift in the model's generative behavior, which is a symptom of factual fragility or reasoning inconsistencies.

\newpage
\section{Benchmark Details}
\label{app:benchmark_details_prompts}

We provide additional details on the six benchmarks used in our evaluation. SVAMP~\citep{patel-etal-2021-nlp} contains arithmetic word problems that require multi-step numerical reasoning and typically produce single numeric answers. MMLU High School Statistics~\citep{hendryckstest2021,hendrycks2021ethics} is an exam-style multiple-choice benchmark that evaluates statistical concepts, conceptual understanding, and quantitative reasoning. CommonsenseQA~\citep{talmor-etal-2019-commonsenseqa} evaluates commonsense and relational inference in a multiple-choice format.

For factuality-oriented evaluation, we use SimpleQA Verified~\citep{haas2025simpleqaverifiedreliablefactuality} and FActScore~\citep{min-etal-2023-factscore}. SimpleQA Verified consists of short-form fact-retrieval questions with verified answers, while FActScore evaluates long-form generations by decomposing outputs into atomic claims and checking whether each claim is supported by external evidence. Finally, HumanEval~\citep{DBLP:journals/corr/abs-2107-03374} evaluates code synthesis through unit tests.

Together, these benchmarks span reasoning, factual recall, long-form factuality, and code generation, covering both constrained answer formats and open-ended generations. We include the prompts used for each benchmark below.

\section{Baseline Details}
\label{app:baseline_details}

Since C3 serves as a proxy for the credibility of model generations, we compare it with related notions of confidence, consistency, and factuality. We include both vanilla black-box confidence estimators and a non-vanilla factuality checking tool.

\paragraph{Vanilla approaches.}
We consider three vanilla black-box baselines. \textbf{Self-reported confidence}~\citep{lin2022teaching} asks the model to output an explicit numeric confidence score, with clearly defined upper and lower bounds, alongside its answer. This baseline tests whether the model can verbalize its own uncertainty in a way that aligns with correctness or factual support. The prompts used for self-reported confidence are provided in Appendix~\ref{app:prompt for self confidence}.

\textbf{Self-consistency}~\citep{DBLP:conf/iclr/0002WSLCNCZ23} estimates confidence from repeated stochastic decoding under the same prompt. We sample $K$ completions and measure agreement among the generated outputs, where higher agreement indicates greater confidence in the model's answer.

\textbf{Paraphrasing consistency}~\citep{portillo-wightman-etal-2023-strength} measures whether model outputs remain stable under meaning-preserving prompt-level paraphrases. We generate $K$ paraphrased variants of the original question and compare the resulting answers using the same consistency framework as self-consistency.

For sampling-based methods, including self-consistency and paraphrasing consistency, we compute agreement over a set of $K$ generated outputs $\mathcal{Y} = \{y_1, y_2, \dots, y_K\}$:
\begin{equation}
\mathcal{C}(\mathcal{Y}) =
\frac{1}{K(K-1)}
\sum_{i=1}^{K}
\sum_{j \neq i}^{K}
k(y_i, y_j),
\label{eq:consistency}
\end{equation}
where $k(y_i, y_j)$ is a similarity function. For fixed-format tasks, we use an indicator function $\mathbf{1}[y_i = y_j]$. For open-ended generation, we use a semantic similarity metric.

\paragraph{Non-vanilla approach.}
We also compare C3 with FActScore~\citep{min-etal-2023-factscore}, a factuality checking tool for long-form generation. FActScore decomposes each output into atomic facts, retrieves evidence from an external knowledge source, and checks whether the evidence supports each claim. In the benchmark setting, Wikipedia is used as the external knowledge source. The final FActScore is computed as the fraction of supported atomic facts.

\newpage

\end{document}